\documentclass[12pt]{article}
\usepackage[margin=1in]{geometry}
\usepackage[natbibapa]{apacite}
\usepackage{amsmath, amsthm, amssymb, amsfonts}
\usepackage{graphicx}
\usepackage{array}
\usepackage{booktabs}
\usepackage{tabularx}
\usepackage{adjustbox}
\usepackage{rotating}
\usepackage{makecell}
\usepackage{subcaption}
\usepackage{setspace}
\usepackage{float}
\usepackage[utf8]{inputenc}
\usepackage[T1]{fontenc}
\usepackage{lmodern}
\usepackage[english]{babel}
\usepackage[dvipsnames,table]{xcolor}
\usepackage{soulutf8}
\usepackage{tikz}
\usetikzlibrary{arrows.meta,positioning,fit,backgrounds,calc,shapes.arrows,decorations.pathreplacing}
\usepackage{pgfplots}
\pgfplotsset{compat=1.17}
\usepackage{listings}
\usepackage{multirow}
\usepackage[switch]{lineno}
\usepackage[hidelinks]{hyperref}

\newcommand{\gd}{DATUM}
\newcommand{\op}[1]{\texttt{#1}}
\newcommand{\outcome}[1]{\textsc{#1}}

\theoremstyle{definition}

\definecolor{stepbad}{RGB}{248,196,196}
\definecolor{stepnext}{RGB}{252,222,170}
\definecolor{stepok}{RGB}{204,232,204}
\definecolor{steptag}{RGB}{110,110,110}
\newcommand{\steptag}[1]{\textsf{\textbf{\color{steptag}\tiny #1}}\,}
\newcommand{\hlbad}[2]{\steptag{#1}{\sethlcolor{stepbad}\hl{#2}}}

\newcommand{\ttbox}[2]{\colorbox{#1}{\kern-2pt\strut#2\kern-2pt}}

\providecommand{\tbl}[2]{\caption{#1}\centering\small\adjustbox{max width=\textwidth}{#2}}

\newcommand{\cts}{\textsf{CTS}}\newcommand{\mia}{\textsf{MIA}}\newcommand{\gre}{\textsf{GRE}}\newcommand{\ibs}{\textsf{IBS}}
\newcommand{\cvt}{\textsf{Converter}}\newcommand{\bnd}{\textsf{Binder}}\newcommand{\tsb}{\textsf{Transcriber}}
\colorlet{wrapperhlc}{orange!8}
\begin{document}

\begin{center}
{\LARGE\bfseries \gd{}: A System of Syntactic Kernel and Semantic Wrapper\\[2pt] for Trustworthy LLM Spatial Reasoning\par}
\vspace{1em}
{\large Zhangyu Wang\par}
\vspace{0.5em}
{\small School of Computing and Information Science, University of Maine, Orono, ME, USA\\
Correspondence: Zhangyu Wang, zhangyu.wang@maine.edu\par}
\end{center}


\begin{abstract}
\noindent AI-generated mathematical proofs can be trusted because formal kernels such as Lean verify them, but spatial reasoning has no such kernel. We present \gd{}, which verifies large language model (LLM) spatial reasoning formally as a typed graph. A semantic wrapper transcribes plain-language question and reasoning into typed spatial statements and splits vague wording such as ``north of'' into explicit interpretation branches. A syntactic kernel checks each statement by fixed rules grounded in GIScience principles and, like a compiler, reports error stacktraces as typed exception reports. \gd{} provides automated, reliable and human-readable verification tools of LLM generated spatial reasoning. On 585 spatial QA tasks, \gd{} detects wrong conclusions with F1 0.95 at 85--87\% coverage, against 0.44--0.81 for existing LLM-as-judge baselines. \gd{}'s exception reports also enables LLM self-correction, fixing 103 out of 124 reasoning errors, against 72--84 for baselines.

\end{abstract}


\section{Introduction}\label{sec:intro}

\begin{figure}[!tp]
\centering
\colorlet{wrapperhl}{orange!8}
\footnotesize
\setlength{\fboxsep}{4pt}
\makebox[\dimexpr0.965\textwidth+2\fboxsep+0.8pt\relax][l]{\hspace{0.8em}\small\bfseries LLM Spatial Reasoning}\\[3pt]
\fbox{\begin{minipage}{0.965\textwidth}
\textbf{Question} (MapEval): What is the direction of the Little Mermaid statue from Copenhagen Central Station?\quad Options: East, West, North, South.\\[2pt]
\textbf{Context}: Little Mermaid (55.6929, 12.5993); Copenhagen Central Station (55.6728, 12.5649).\\[2pt]
\textbf{Reasoning} (GPT-4o, abridged): The latitude of the Little Mermaid statue (55.6929) is greater \ldots\ the statue is to the north \ldots\ The longitude \ldots\ (12.5993) is greater \ldots\ the statue is to the east \ldots\ {\sethlcolor{wrapperhl}\hl{Since the Little Mermaid statue is both north and east of Copenhagen Central Station, the primary direction considering the options given is \textbf{North}.}}\\[2pt]
\textbf{Conclusion statement}: North.
\end{minipage}}\\[0pt]
\begin{tikzpicture}[font=\footnotesize,
  box/.style={draw,rounded corners=2pt,align=center,inner sep=4pt},
  arr/.style={-{Stealth[length=2mm]},thick,gray!70!black},
  nsec/.style={fill=blue!8}, esec/.style={fill=violet!10},
  tick/.style={font=\scriptsize,inner sep=1pt,gray!40!black},
  info/.style={anchor=west,align=left,text width=7.0cm}]
\begin{scope}[shift={(4.9,4.0)},x=0.857cm,y=0.857cm]
\fill[nsec] (0,0) -- (0,2.8) -- (2.8,2.8) -- cycle;
\fill[esec] (0,0) -- (2.8,2.8) -- (2.8,0) -- cycle;
\draw[gray,dashed] (0,0) -- (2.8,2.8);
\draw[gray!60] (0,0) rectangle (2.8,2.8);
\foreach \t in {1,2} {\draw[gray!60] (\t,0) -- (\t,-0.07) node[below,tick] {\t}; \draw[gray!60] (0,\t) -- (-0.07,\t) node[left,tick] {\t};}
\node[tick,anchor=north] at (1.4,-0.32) {km east};
\node[tick,anchor=south,rotate=90] at (-0.36,1.4) {km north};
\node[blue!60!black] at (0.75,2.4) {North};
\node[violet!70!black] at (2.2,0.5) {East};
\draw[-{Stealth[length=1.8mm]},thick] (0,0) -- (2.16,2.22);
\fill[red!70!black] (2.16,2.22) circle[radius=0.07];
\fill (0,0) circle[radius=0.07];
\draw[gray!70!black] (0,0) ++(90:0.8) arc[start angle=90,end angle=46,radius=0.8];
\node[anchor=west,inner sep=1pt,font=\scriptsize] at (0.1,1.0) {44.0$^\circ$};
\coordinate (m1a) at (-1.11,-0.75); \coordinate (m1b) at (2.8,2.8); \coordinate (m1r) at (2.8,1.4);
\end{scope}
\node[info] (i1) at ($(m1r)+(0.35,0)$) {\textbf{Interpretation 1:}\\\textbf{compass bearing}\\ drawn on a local grid, in kilometres\\[5pt] bearing 44.0$^\circ$ lies in the North sector\\ $\Rightarrow$ North \quad \textcolor{green!40!black}{\textbf{conclusion holds}}};
\node[fit={(m1a)(m1b)(i1)},draw=gray!60,rounded corners=2pt,inner sep=3pt] (k1) {};
\begin{scope}[shift={(4.9,0)},x=0.686cm,y=0.686cm]
\fill[nsec] (0,0) -- (0,3.5) -- (3.5,3.5) -- cycle;
\fill[esec] (0,0) -- (3.5,3.5) -- (3.5,0) -- cycle;
\draw[gray,dashed] (0,0) -- (3.5,3.5);
\draw[gray!60] (0,0) rectangle (3.5,3.5);
\foreach \t in {1,2,3} {\draw[gray!60] (\t,0) -- (\t,-0.09) node[below,tick] {.0\t}; \draw[gray!60] (0,\t) -- (-0.09,\t) node[left,tick] {.0\t};}
\node[tick,anchor=north] at (1.75,-0.4) {$\Delta$lon (degrees)};
\node[tick,anchor=south,rotate=90] at (-0.75,1.75) {$\Delta$lat (degrees)};
\node[blue!60!black] at (0.95,3.0) {North};
\node[violet!70!black] at (2.75,0.62) {East};
\draw[gray!70!black,densely dotted] (3.44,0) -- (3.44,2.01) -- (0,2.01);
\draw[-{Stealth[length=1.8mm]},thick] (0,0) -- (3.44,2.01);
\fill[red!70!black] (3.44,2.01) circle[radius=0.09];
\fill (0,0) circle[radius=0.09];
\coordinate (m2a) at (-1.38,-0.95); \coordinate (m2b) at (3.5,3.5); \coordinate (m2r) at (3.5,1.75);
\end{scope}
\node[info] (i2) at ($(m2r)+(0.35,0)$) {\textbf{Interpretation 2:}\\\textbf{larger coordinate difference}\\ drawn on a plate carr\'ee, in degrees\\[5pt] $|\Delta\text{lon}|$ 0.0344$^\circ$ $>$ $|\Delta\text{lat}|$ 0.0201$^\circ$\\ $\Rightarrow$ East \quad \textcolor{red!70!black}{\textbf{conclusion fails}}};
\node[fit={(m2a)(m2b)(i2)(m2a -| i1.east)},draw=gray!60,rounded corners=2pt,inner sep=3pt] (k2) {};
\coordinate (wm) at ($(k1.south)!0.5!(k2.north)$); \coordinate (wx) at (1.65,0);
\node[box,fill=wrapperhl,draw=orange!60!black,text width=3.1cm] (w) at (wx |- wm) {``the primary direction''\\ admits two interpretations};
\draw[arr] (w.east) -- (k1.west);
\draw[arr] (w.east) -- (k2.west);
\node[font=\small\bfseries,anchor=south] (kt) at ($(k1.north)+(0,1mm)$) {Syntactic Kernel};
\node[font=\small\bfseries,anchor=base] at (w.north |- kt.base) {Semantic Wrapper};
\node[font=\footnotesize\itshape,gray!30!black,anchor=north] (kb) at ($(k2.south)+(0,-1mm)$) {checking individual branches};
\node[font=\footnotesize\itshape,gray!30!black,anchor=base] (wb) at (w.south |- kb.base) {interpreting ambiguity};
\begin{scope}[on background layer]
\node[fit={($(kt.north)+(0,2.5mm)$)(kb)(wb)(k1)(k2)(w)},draw=black!70,line width=0.8pt,fill=gray!4,rounded corners=4pt,inner sep=6pt] (D) {};
\end{scope}
\node[fill=white,inner sep=2pt,anchor=west,font=\large\bfseries] at ($(D.north west)+(4mm,0)$) {\gd{}};
\node[single arrow,draw=none,fill=gray!45,rotate=-90,minimum height=5.5mm,minimum width=10mm,single arrow head extend=2.5mm,anchor=tip] at (D.north) {};
\node[single arrow,draw=none,fill=gray!45,rotate=-90,minimum height=5.5mm,minimum width=10mm,single arrow head extend=2.5mm,anchor=tail] at (D.south) {};
\end{tikzpicture}\\[0pt]
\makebox[\dimexpr0.965\textwidth+2\fboxsep+0.8pt\relax][l]{\hspace{0.8em}\small\bfseries Verification Report}\\[3pt]
\fbox{\begin{minipage}{0.965\textwidth}
\setlength{\fboxsep}{1pt}%
\scriptsize\ttfamily\raggedright
\textbf{\gd{}}: \outcome{\rmfamily Contingent}\\[2pt]
mapeval/557: conditionally verified [GD-C101]: the conclusion agrees only under some interpretations\\
\kern-\fboxsep\colorbox{wrapperhl}{\renewcommand{\arraystretch}{0.9}\begin{tabular}{@{}l@{}}ReadingSplit: the conclusion depends on the interpretation\\
\hspace*{2em}holds under direction=bearing\symbol{95}sector4\\
\hspace*{2em}fails under direction=dominant\symbol{95}axis\end{tabular}}\\
OK: 4 of 7 transcribed steps agree with the map context\\
\textbf{Fix}: state which interpretation the question means; the conclusion holds only under those listed
\end{minipage}}
\normalsize
\caption{A MapEval question checked by \gd{}. Top: the question, its context and the GPT-4o reasoning. Middle: \gd{} takes in the reasoning; its semantic wrapper instantiates two branches, one per interpretation of ``the primary direction'', and its syntactic kernel checks the reasoning under each interpretation separately, drawn on the map projection in which that interpretation is natural. Bottom: the verification report \gd{} produces, in the style of a Java exception. The conclusion statement is neither right nor wrong until the interpretation is fixed. The shaded sentence, the wrapper box and the shaded report lines mark the same ambiguity as it passes through the system.}
\label{fig:opening}
\end{figure}

Geographic decisions are increasingly made with the help of AI. Large language models (LLMs) find the nearest pharmacy in a list of places, plan trips under opening hours, judge whether one landmark lies north of another, and propose the steps of GIS workflows \citep{li2023autonomous,akinboyewa2025giscopilot,dihan2024mapeval}. However, AI cannot yet be fully trusted in geographical applications, due to the existence of spatial ambiguity and the lack of geospatial world knowledge. Recent GIScience work has summarized common LLM failures into three categories of \emph{geospatial hallucinations}: spatial ignorance, spatial vagueness and spatial reasoning inconsistency \citep{liu2026spatialcot}.

The problem grows as GIS becomes agentic. Autonomous GIS systems interpret a request, plan a workflow, select data and tools, and execute the analysis with little human involvement \citep{li2023autonomous,akinboyewa2025giscopilot}. Every step consumes the output of the step before it. A pipeline that is not monitored and checked can therefore hallucinate a value, carry the error into every later step, where it accumulates, and break on its own, with nothing in its final output to show where it went wrong. The remedy is an ``inspector'': a verification system that checks every step (its inputs, transformations and outputs), raises an alarm whenever errors occur, and locates the causes so that users can fix them. In mathematics, such an inspector already exists: AI-generated proofs are checked by a formal kernel such as Lean \citep{demoura2021lean4}. For geography, no such verification system exists yet.

Building a verification system for geography is even harder than in mathematics. The difficulty lies in the semantic vagueness of statements being checked. A mathematical statement has one fixed meaning; a geographic one does not necessarily. ``North of'' can mean a compass sector or a dominant axis. ``At'' a place can mean a point location or a polygon region. A distance written as ``3.6 km'' is a measurement whose accuracy depends on how it was recorded. A listing of places may be incomplete, and what counts as true can depend on place and community \citep{janowicz2025whosetruth}. If the verification system itself had to decide what such words mean, it could pick one meaning without telling the user, and possibly check a statement the question never meant.

We therefore propose to address the problem by separating semantic vagueness from syntactic soundness in verification. A \emph{syntactic kernel} checks precise, typed spatial statements against data by fixed rules, grounded in what GIScience has long established, such as the core concepts of spatial information \citep{kuhn2012core}, topological relations \citep{egenhofer1991point} and the accuracy of spatial data \citep{duckham2001formal}. A \emph{semantic wrapper} turns words into such statements and makes every choice of meaning explicit, listing each admissible interpretation of vague language \citep{burrough1996indeterminate,bennett2010spatial}. The kernel stays small and trustworthy, and the ambiguity of geographic language is handled as explicitly announced and checked conditions. 

Based on the distinction between a syntactic kernel and a semantic wrapper, we designed a geographic verification system called \gd{} (\textbf{D}eterministic \textbf{A}ssessment of \textbf{T}yped \textbf{U}tterances and \textbf{M}easurements). Like a geodetic datum, the fixed reference against which positions are measured, \gd{} gives LLM spatial reasoning a fixed reference to be checked against. It first reads a spatial question, the context (e.g. a map) supplied with it, and an LLM's reasoning; second, its wrapper transcribes plain language descriptions into a computable reasoning graph; third, its kernel runs formal checks on the reasoning graph and locates errors or reasoning gaps; finally, a human-readable report is assembled. \gd{} fills a gap left by two lines of existing work: generation-side work tries to improve how a model reasons or writes code, for example with core geospatial concepts \citep{liu2026spatialcot,wang2026coreconcepts}; input-side tools certify the quality and provenance of the data an agent reads \citep{cabrera2026geoproof,mapsmith2026}; neither checks the \emph{inference}: whether a given piece of spatial reasoning over the supplied data is right, step by step, in a way a GIS user can verify.

We closely incorporated GIScience domain knowledge into the implementation of \gd{}. The kernel defines types built on the core concepts of spatial information \citep{kuhn2012core}. Each verdict is derived mechanically from a fixed set of typed geographical inference rules, so its soundness can be formally checked by a geographer, and when new inference rules are needed, geographers can manually expand the kernel. The wrapper uses the knowledge of LLMs for what rule-based engines cannot handle, such as interpreting vague wording and free-form reasoning, and turns them into typed statements. Where the wording is ambiguous, the wrapper instantiates one branch per interpretation, and the kernel checks the reasoning under each branch separately (Section~\ref{sec:design-wrapper}). A typed contract interface between the wrapper and the kernel screens out malformed, ill-typed or unsupported statements and reports them before they enter the kernel.

Figure~\ref{fig:opening} depicts a concrete example of how DATUM detects errors from an LLM-generated reasoning for Question 557 in MapEval, a benchmark of map-based geospatial questions \citep{dihan2024mapeval}. The question asks: \emph{What is the direction of the Little Mermaid statue from Copenhagen Central Station?} The options are East, West, North and South, and the question supplies the coordinates of both places. A GPT-4o reasoning correctly finds that the statue lies both north and east of the station, but hastily concludes that ``the primary direction considering the options given is North''. \gd{}'s wrapper finds that ``the primary direction'' has two ambiguous meanings, and the reasoning branches based on different meanings disagree on which conclusion is correct: if ``the primary direction'' means the compass bearing on a local metric grid, the statue lies at $44.0^\circ$, just inside the North sector, so the correct conclusion is ``North''; if ``the primary direction'' means the larger of the two coordinate differences, as it appears on a plate carr\'ee map of raw longitude and latitude, the longitude difference in degrees exceeds the latitude difference, so the correct conclusion is ``East''. \gd{} successfully yields a \outcome{Contingent} verdict, asserting that the LLM's conclusion statement ``North'' only holds under the compass bearing interpretation and fails under the coordinate-difference interpretation. LLM-as-judge baselines, on the contrary, disagree on whether ``North'' is the correct conclusion, and none recognizes that the dispute comes from an ambiguous spatial interpretation.

We do notice that the LLM use in the wrapper still poses some risk of untrustworthy verifications, e.g. an LLM may hallucinate a non-stated approximation rule such as ``use Euclidean distance instead of haversine distance''. However, our wrapper-contract-kernel design makes sure that all LLM-transcribed statements and LLM-generated interpretations are recorded as explicit branches that a person can verify, so that even if the wrapper makes mistakes, they can be immediately traced and properly solved.

Our contributions are:
\begin{enumerate}
\item \textbf{A syntactic--semantic architecture for verifying spatial reasoning.} \gd{} separates semantic interpretation, which maps natural-language questions and reasoning to typed spatial statements, from the syntactic verification of those statements. A semantic wrapper performs the first, a syntactic kernel the second, and typed contracts govern the interface between them. The LLM is confined to the wrapper, and ambiguity in the wording becomes explicit branches rather than hidden choices (Section~\ref{sec:split}).
\item \textbf{A typed spatial kernel.} The kernel is a formal system whose operations carry core-concept type signatures \citep{scheider2020ontology}. It runs three independent checks: it recomputes the conclusion under every interpretation supplied by the wrapper; it checks each step of the reasoning by its typed operation; and it asks whether the steps it confirmed entail the conclusion statement. Measurements are compared within their declared accuracy, and spatial relations are checked against reference gazetteers and geometry databases (Section~\ref{sec:method-kernel}).
\item \textbf{An LLM-assisted semantic wrapper.} A deterministic converter extracts structured inputs such as place names, addresses and coordinates into typed records of values, keeping the written precision of every number. Two LLM assistants, a \emph{binder} and a \emph{transcriber}, turn plain language statements and free-form values into typed statements under strict schemas, each quoting its source verbatim. Vague wording is declared as interpretations, each instantiated as a kernel branch, and the \outcome{Contingent} outcome reports the verdict under every interpretation, following supervaluation semantics \citep{fine1975vagueness} and the place-dependence of geographic truth \citep{janowicz2025whosetruth} (Sections~\ref{sec:method-records} to~\ref{sec:method-contracts}).
\item \textbf{A typed verification report in the style of Java exceptions.} Failed checks raise exceptions from a hierarchy of 22 classes in 8 families, which, as in Java, can be caught at any level. The report names the exception at the first wrong step, traces it through each later step to the conclusion statement as a stack trace, lists other wrong steps as warnings, and ends with a fix, all deterministically rendered from the kernel's typed states. The hierarchy also separates the three kinds of geospatial hallucinations into typed errors with different fixes (Sections~\ref{sec:method-exceptions} and~\ref{sec:res-kinds}).
\item \textbf{Quantitative gains over LLM-as-judge and re-solving baselines.} We give \gd{} 585 questions from MapEval and GS-QA together with the reasonings generated by GPT-4o, and ask \gd{} to verify the validity of these reasonings. \gd{} gives a verdict on 87\% of them (its coverage), with an F1 of 0.95 on the covered cases in detecting wrong conclusions, 0.95--0.96 when any of five other models from three providers transcribes the reasoning, and, when the model also states the question, 0.95 on Gemini 4 Argon and 0.86--0.89 on the other four, against 0.44 to 0.81 for baselines on the same models, including baselines that re-solve the task and decide only where the re-solve and a judge agree. The remaining 13\% are left undecided, each with a stated reason, such as evidence missing from the supplied data. Its reports also let the LLM repair 103 of 124 wrong conclusions, and it found an error in a published ground-truth label (Section~\ref{sec:setup} and Appendix~\ref{sec:cases}).
\end{enumerate}

\section{Background and related work}\label{sec:related}

\subsection{Core concepts and concept transformations}
\citet{kuhn2012core} proposed a small set of core concepts of spatial information (location, neighbourhood, field, object, network and event, together with the quality concepts granularity and accuracy) as a common vocabulary across disciplines. Later work built on them a language for spatial computing \citep{kuhn2015designing}, core concept data types \citep{scheider2020ontology}, an algebra of concept transformations \citep{steenbergen2023procedural}, and a grammar that reads a geo-analytical question as a transformation from input concepts to an output concept \citep{xu2023grammar,nyamsuren2023semantic}. The common idea is that a spatial question is a \emph{typed} transformation. We adopt this idea and add what the line has not yet needed: types that are \emph{checked against data}, with a defined error for each check that fails.

\subsection{LLM spatial reasoning and geospatial hallucination}
Benchmarks such as MapEval \citep{dihan2024mapeval}, GeoBenchX \citep{krechetova2025geobenchx} and others \citep{ji2025foundation,yang2025spatialcognition} show that LLMs still fail at distance, direction, routing and planning tasks. GIS agents \citep{li2023autonomous,akinboyewa2025giscopilot,zhang2024geogpt,wei2025geotoolgpt} and spatial retrieval \citep{yu2025spatialrag,liang2025geographrag} improve generation. Spatial CoT \citep{liu2026spatialcot} guides the reasoning of a model with core concepts and transformation graphs \citep[see also][]{xu2023grammar}, and names three kinds of geospatial hallucination. In the same generation-side spirit, prompts enriched with core geospatial concepts improve the quality of LLM-generated geospatial code \citep{wang2026coreconcepts}. Its own analysis notes that exact geometric work may need execution-based verification, and it scores hallucinations with an LLM judge; \gd{} computes them.

\subsection{Certifying the inputs of AI GIS}
Practitioners have started to certify the \emph{inputs} of AI GIS systems. GeoProof, an open-source server between AI agents and spatial data, attaches ratings of accuracy, completeness, freshness, provenance and authority to the data an agent reads \citep{cabrera2026geoproof}, and MapSmith, a geoprocessing server for agents, computes all geometry with deterministic tools and records a provenance manifest for each output \citep{mapsmith2026}. MapSmith also validates an agent's plan, but only structurally: by its own account, a well-formed plan that addresses the wrong question still runs. These tools tell a user whether the data can be trusted; \gd{} checks what is inferred from the data, and the two are complementary.

\subsection{Verifying reasoning}
Hoare logic \citep{hoare1969axiomatic} checks a program step by step against pre- and postconditions; HoarePrompt \citep{bouras2026hoareprompt} applies the same idea to LLM reasoning about code. Proof assistants keep the trusted part small: the de Bruijn criterion asks that a proof be checkable by a simple kernel, however it was produced, and Lean separates an untrusted elaborator from a small kernel \citep{demoura2021lean4}. Execution-based checks \citep{shi2022execution} and LLM judges, which are sensitive to position, verbosity and self-preference \citep{zheng2023judging}, are the common alternatives. In mainstream LLM research, hallucination detection checks generated claims by sampling consistency \citep{manakul2023selfcheckgpt}, retrieved evidence \citep{min2023factscore} or self-posed verification questions \citep{dhuliawala2024cove}, and surveys separate unfaithfulness to the given context from factual error \citep{ji2023survey}. A spatial claim that contradicts its map context is an unfaithfulness hallucination, and \gd{} detects it by recomputation rather than by sampling or retrieval. \gd{} follows the proof-assistant design: the LLM elaborates, and the kernel checks.

\subsection{Vagueness, uncertainty and place-dependent truth}
GIScience has long treated vague regions and relations formally: indeterminate boundaries \citep{burrough1996indeterminate}, egg-yolk regions \citep{cohn1996eggyolk}, spatial vagueness \citep{bennett2010spatial} and imperfection in geographic information \citep{duckham2001formal}. Supervaluation semantics \citep{fine1975vagueness} calls a vague statement true if it is true under every admissible sharpening, false if false under every one, and otherwise neither. Janowicz and colleagues argue that what counts as true or appropriate can itself depend on place and community \citep{janowicz2025whosetruth,janowicz2018debiasing}. Our \outcome{Contingent} outcome is the supervaluationist ``neither'' made explicit: the interpretations, including those that carry a geographic convention, are listed with their verdicts.

\begin{table}
\tbl{Related work compared by what it checks. \checkmark: yes; partly: an LLM judge reads the reasoning but does not check it step by step; --: no.}
{\begin{tabular}{lccccc}
\toprule
Work & \makecell{Checks\\reasoning steps} & \makecell{Deterministic\\verdict} & \makecell{Typed\\errors} & \makecell{Vague\\interpretations} & \makecell{GIScience\\grounding} \\
\midrule
Autonomous GIS \citep{li2023autonomous} & -- & -- & -- & -- & -- \\
GIS Copilot \citep{akinboyewa2025giscopilot} & -- & -- & -- & -- & -- \\
Spatial CoT \citep{liu2026spatialcot} & -- & -- & -- & -- & \checkmark \\
Core-concept prompts \citep{wang2026coreconcepts} & -- & -- & -- & -- & \checkmark \\
\midrule
LLM-as-judge \citep{zheng2023judging} & partly & -- & -- & -- & -- \\
HoarePrompt \citep{bouras2026hoareprompt} & \checkmark & -- & -- & -- & -- \\
\midrule
\gd{} (ours) & \checkmark & \checkmark & \checkmark & \checkmark & \checkmark \\
\bottomrule
\end{tabular}}
\label{tab:related}
\end{table}

\section{Design Principles of DATUM}\label{sec:split}\label{sec:principles}

\begin{figure}[!tp]
\centering
\includegraphics[width=\textwidth]{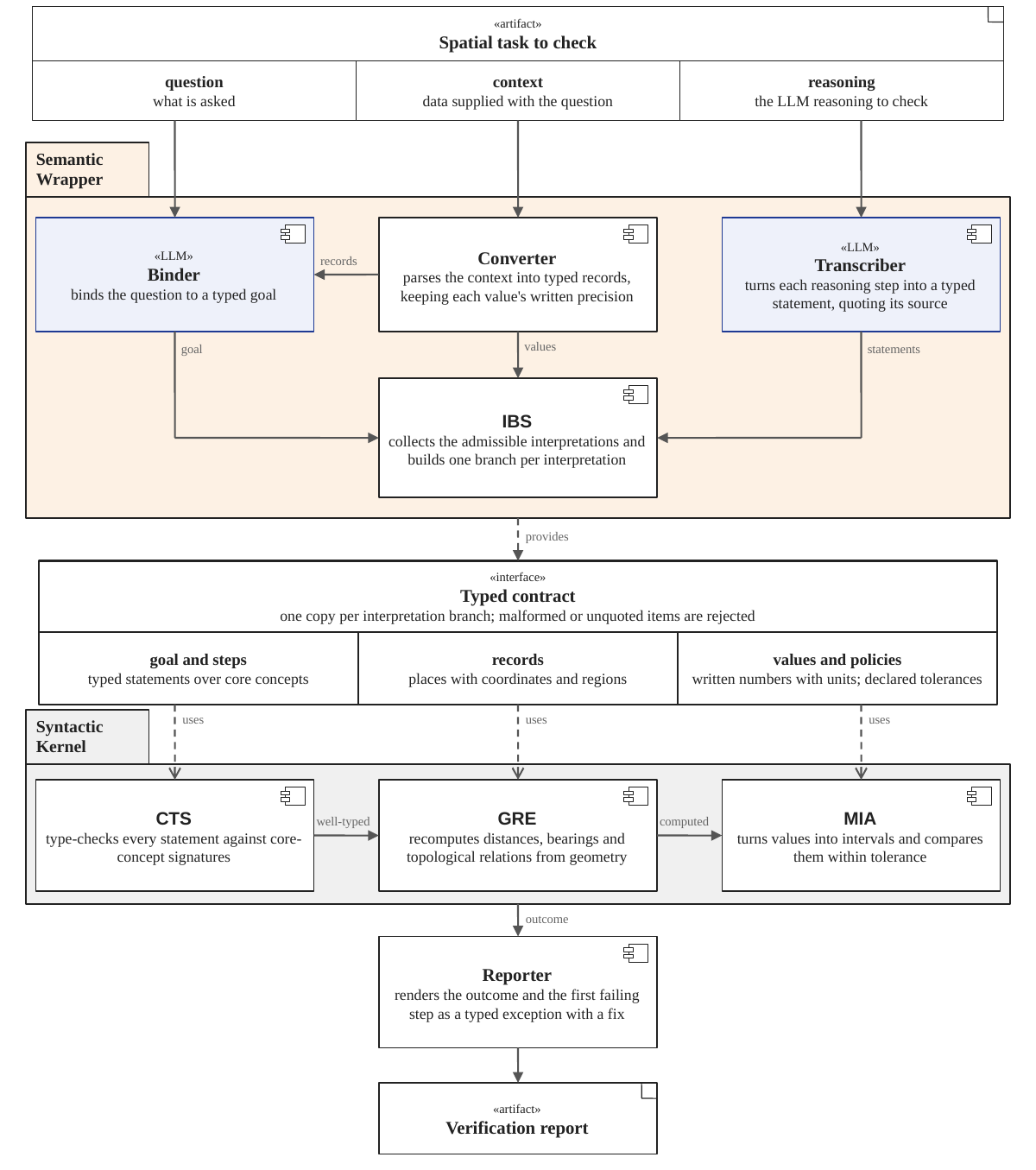}
\caption{Architecture of \gd{} as a UML component diagram. The semantic wrapper turns the three parts of the spatial task into the typed contract: the Binder reads the question, the Converter the context and the Transcriber the reasoning, and all three pass their interpretations to \ibs{}, which builds one copy of the contract per interpretation branch. Each part of the contract is used by one kernel component: \cts{} type-checks the goal and steps, \gre{} computes spatial relations from the records, and \mia{} compares the claimed values with the computed ones within tolerance. The reporter renders the outcome as the verification report. Blue-shaded components are LLM-assisted. Others are deterministic.}
\label{fig:architecture}
\end{figure}

\gd{} consists of two halves, a \emph{Semantic Wrapper} (in short, \emph{wrapper}) and a \emph{Syntactic Kernel} (in short, \emph{kernel}), joined by \emph{Typed Contracts} (in short, \emph{contracts}) (Figure~\ref{fig:architecture}). This section follows Figure~\ref{fig:architecture} from top to bottom: the role of the wrapper, the contract it hands over, the kernel that checks it, and the outputs a user sees; Section~\ref{sec:method} describes how each component is implemented.

\subsection{Semantic Wrapper}\label{sec:design-wrapper}
The Semantic Wrapper is the front end of \gd{}. Its input is the raw natural-language material of a spatial task: the question (e.g. finding the nearest restaurant), the context (e.g. maps, lists of candidate locations, opening hours), and the reasoning text (e.g. an LLM-generated reasoning), which ends with its \emph{conclusion statement}, the final claim the reasoning offers as the solution to the question. The raw input is fed through four modules with different degrees of trust: a deterministic \cvt{} parses formatted data (e.g. YYYY-MM-DD timestamp strings) into typed records; an LLM-assisted \bnd{} reads the question and binds it to a typed goal (e.g. a \op{geometry} goal for the distance to the nearest restaurant; Table~\ref{tab:goals}); an LLM-assisted \tsb{} transcribes the reasoning text into typed statements; a deterministic \ibs{} (Interpretation Branch Splitter) module collects semantically ambiguous interpretations of goals (e.g. what ``North'' means), values (e.g. 10:00 without AM or PM) and statements (e.g. in Georgia -- state or country?) from the other three components and instantiate a separate verification branch for each interpretation. Its output is a set of typed objects acceptable to the Syntactic Kernel: a typed goal, typed records and typed reasoning statements (Section~\ref{sec:design-kernel}).

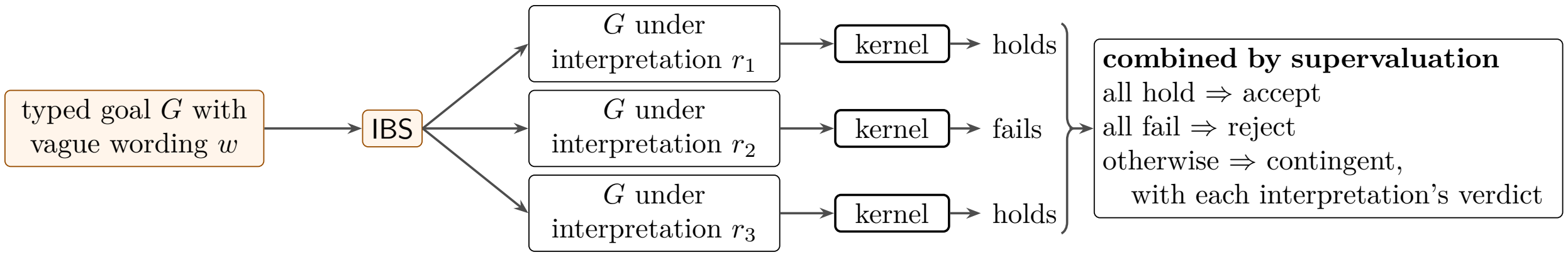
\begin{figure}[t]
\centering
\adjustbox{max width=\textwidth}{
\begin{tikzpicture}[font=\footnotesize,
  box/.style={draw,rounded corners=2pt,align=center,inner sep=3pt},
  arr/.style={-{Stealth[length=1.8mm]},thick,gray!60!black}]
\node[box,fill=wrapperhlc,draw=orange!60!black,text width=3.0cm] (g) at (0,0) {typed goal $G$ with vague wording $w$};
\node[box,draw=orange!60!black,fill=wrapperhlc] (ibs) at (3.2,0) {\ibs{}};
\node[box,text width=2.9cm] (b1) at (6.45,1.05) {$G$ under\\interpretation $r_1$};
\node[box,text width=2.9cm] (b2) at (6.45,0) {$G$ under\\interpretation $r_2$};
\node[box,text width=2.9cm] (b3) at (6.45,-1.05) {$G$ under\\interpretation $r_3$};
\node[box,thick,text width=1.2cm] (k1) at (9.4,1.05) {kernel};
\node[box,thick,text width=1.2cm] (k2) at (9.4,0) {kernel};
\node[box,thick,text width=1.2cm] (k3) at (9.4,-1.05) {kernel};
\node[anchor=west] (v1) at (10.5,1.05) {holds};
\node[anchor=west] (v2) at (10.5,0) {fails};
\node[anchor=west] (v3) at (10.5,-1.05) {holds};
\node[box,text width=5.6cm,align=left,anchor=west] (c) at (11.9,0) {\textbf{combined by supervaluation}\\ all hold $\Rightarrow$ accept\\ all fail $\Rightarrow$ reject\\ otherwise $\Rightarrow$ contingent,\\ \hspace*{1em}with each interpretation\textquoteright s verdict};
\draw[arr] (g) -- (ibs);
\foreach \i in {1,2,3} {\draw[arr] (ibs.east) -- (b\i.west); \draw[arr] (b\i) -- (k\i); \draw[arr] (k\i.east) -- (v\i.west);}
\draw[decorate,decoration={brace,amplitude=4pt},thick,gray!60!black] (11.5,1.3) -- (11.5,-1.3);
\draw[arr] (11.6,0) -- (c.west);
\end{tikzpicture}}
\caption{\ibs{} in the wrapper. Vague wording is split into one branch per admissible interpretation; the kernel checks each branch independently, and the verdicts are combined by supervaluation. Figure~\ref{fig:opening} shows a two-branch case.}
\label{fig:ibs}
\end{figure}

\ibs{} bears the central obligation of handling spatial semantic ambiguity (Figure~\ref{fig:ibs}). Which interpretations are admissible is not decided case by case: for each goal type, \gd{} keeps a declared table of interpretation axes (for example, compass sector or dominant axis for a direction, and city or country for ``at'' a place), and \cvt{} flags values written with two meanings, such as a clock time without AM or PM. \ibs{} collects every axis that applies to the case and assembles one set of typed goal, typed records and typed statements for each combination, called a \emph{branch}. Then, each branch is passed to the Syntactic Kernel for formal verification. When there are multiple interpretations, a reasoning is no longer decisively correct or wrong: a reasoning that holds under every interpretation is correct; a reasoning that fails under every interpretation is wrong; a reasoning that holds under some interpretations and fails under others is contingent, following the conventions of supervaluation semantics \citep{fine1975vagueness} and the place-dependence of geographic truth \citep{janowicz2025whosetruth}. Coverage is bounded by what has been declared: an ambiguity that no axis names passes unnoticed, as in the Erd\H{o}s case \citep{erdos728writeup}, where a kernel confirmed a statement that was not the one the question meant. Because the axis tables are data rather than code, geographers can extend them as new ambiguities are found. Tolerances are declared in the same way, as policies of the task rather than choices made case by case. In this paper each dataset declares one policy (Section~\ref{sec:design-kernel}); the architecture also admits several, for example when a reasoning rounds its values and the question states no precision, in which case each policy becomes a branch and a reasoning that passes under only some of them is reported as contingent.

\subsection{Typed Contracts}\label{sec:design-contract}
The \bnd{} and the \tsb{} use LLMs to help parse values and transcribe statements. To ensure that the typed content generated by LLMs are acceptable to the Syntactic Kernel, we enforce a typed contract check on the Semantic Wrapper output before it reaches the kernel. Each contract is a fixed, deterministic test on one kind of wrapper output \citep{meyer1992contract}. For example, a typed statement must quote the exact words it was transcribed from, so that every statement verdict can be traced back to a span of the original reasoning; a claimed number must appear in that quote, so that a value cannot drift between the reasoning and its typed form; every operation must be a supported one in the kernel and carries the correct signature. A statement that quotes a place the context does not include is marked as uncheckable, i.e. necessary information is not provided to \gd{}.

Contracts check form, not meaning. A binding that is well formed, correctly quoted and well typed, could still contain hallucinated place names. This happened once in our evaluation (MapEval 58, Appendix~\ref{sec:cases}). What the contracts guarantee is that such an error is visible: if a wrong reasoning passed the Typed Contracts and the Syntactic Kernel, the error must have appeared during the LLM binding and transcribing. The contracts grow with the kernel -- if the Syntactic Kernel accepts a new type of reasoning verification (e.g. 3D map planning), the corresponding contract needs to be added to the Typed Contracts.

\subsection{Syntactic Kernel}\label{sec:design-kernel}
The Syntactic Kernel is the trusted core of \gd{}. Its input is the typed content (goals, records, statements) produced by the Semantic Wrapper that pass the contracts. Its output is a verdict on each statement within each branch: valid, invalid, or uncheckable when the data cannot settle it. It is \emph{syntactic} in the logician's sense: its rules look only at the form of a statement and the values in the records, never at the original words that produced them.

\begin{figure}[t]
\centering
\tikzset{
  kc/.style={font=\scriptsize},
  kbox/.style={use as bounding box},
  loc/.style={draw=blue!60!black,fill=blue!8,rounded corners=2pt,inner sep=2.5pt,minimum height=4.5mm},
  obj/.style={draw=green!45!black,fill=green!8,rounded corners=2pt,inner sep=2.5pt,minimum height=4.5mm},
  fld/.style={draw=orange!70!black,fill=orange!10,rounded corners=2pt,inner sep=2.5pt,minimum height=4.5mm},
  amt/.style={draw=gray!70!black,fill=gray!10,rounded corners=2pt,inner sep=2.5pt,minimum height=4.5mm},
  op/.style={draw,very thick,rounded corners=1pt,inner sep=2.5pt,minimum height=4.5mm,font=\scriptsize\ttfamily},
  karr/.style={-{Stealth[length=1.5mm]},thick,gray!60!black}}
\providecommand{\kcbox}{}\renewcommand{\kcbox}{\path[kbox] (-0.1,-0.75) rectangle (5.0,3.35);}
\begin{minipage}[t]{0.32\textwidth}\centering
{\footnotesize\bfseries (a) \cts{}: concept types}\\[6pt]
\begin{tikzpicture}[kc]
\kcbox
\node[loc] (l1) at (0.65,3.0) {Location};
\node[loc] (l2) at (0.65,2.35) {Location};
\node[op] (d) at (2.55,2.675) {distance};
\node[amt,anchor=west] (a1) at (3.75,2.675) {Amount};
\draw[karr] (l1.east) -- (d.west); \draw[karr] (l2.east) -- (d.west); \draw[karr] (d.east) -- (a1.west);
\node[obj] (s1) at (0.65,1.55) {Set$\langle$Object$\rangle$};
\node[fld] (f1) at (0.65,0.9) {Field};
\node[op] (n) at (2.55,1.225) {nearest};
\node[obj,anchor=west] (o1) at (3.75,1.225) {Object};
\draw[karr] (s1.east) -- (n.west); \draw[karr] (f1.east) -- (n.west); \draw[karr] (n.east) -- (o1.west);
\node[obj] (s2) at (0.65,0.1) {Set$\langle$Object$\rangle$};
\node[loc] (l3) at (0.65,-0.55) {Location};
\node[op,draw=red!70!black] (d2) at (2.55,-0.225) {distance};
\draw[karr] (s2.east) -- (d2.west); \draw[karr] (l3.east) -- (d2.west);
\node[text=red!70!black,align=left,anchor=west] at (3.3,-0.225) {\textbf{$\times$} ill-typed,\\\phantom{$\times$} rejected};
\end{tikzpicture}
\end{minipage}\hfill
\begin{minipage}[t]{0.32\textwidth}\centering
{\footnotesize\bfseries (b) \mia{}: numerical intervals}\\[6pt]
\begin{tikzpicture}[kc]
\kcbox
\fill[green!15] (1.1,0) rectangle (3.1,2.75);
\draw[green!45!black,densely dashed] (1.1,0) -- (1.1,2.75) (3.1,0) -- (3.1,2.75);
\node[green!35!black,align=center] at (2.1,3.05) {tolerance band};
\draw[blue!60!black,very thick] (2.1,0) -- (2.1,1.25);
\node[blue!60!black,anchor=south,inner sep=2pt] at (2.1,1.25) {computed};
\draw[black,thick,{Bar[width=2.2mm]}-{Bar[width=2.2mm]}] (2.3,2.2) -- (2.8,2.2);
\node[green!35!black,anchor=south,inner sep=3pt] at (2.55,2.2) {\textbf{\checkmark} agrees};
\draw[red!70!black,thick,{Bar[width=2.2mm]}-{Bar[width=2.2mm]}] (3.55,1.0) -- (4.05,1.0);
\node[red!70!black,anchor=south,inner sep=3pt] at (3.8,1.0) {\textbf{$\times$} refuted};
\draw[gray!70!black,-{Stealth[length=1.3mm]}] (0,0) -- (4.85,0);
\foreach \v/\l in {1.2/3,3.2/4} {\draw[gray!70!black] (\v,-0.06) -- (\v,0.06); \node[below,inner sep=2pt] at (\v,-0.06) {\l};}
\node[below,inner sep=2pt] at (4.45,-0.06) {km};
\node[anchor=west,inner sep=1pt,gray!40!black] at (0.0,-0.55) {\tikz[baseline=-0.6ex]\draw[thick,{Bar[width=1.8mm]}-{Bar[width=1.8mm]}] (0,0) -- (0.35,0); claim, as precise as its digits};
\end{tikzpicture}
\end{minipage}\hfill
\begin{minipage}[t]{0.32\textwidth}\centering
{\footnotesize\bfseries (c) \gre{}: spatial relations}\\[6pt]
\begin{tikzpicture}[kc]
\kcbox
\fill[orange!12] (2.2,1.3) ellipse[x radius=1.85cm,y radius=1.25cm];
\draw[orange!70!black,densely dashed] (2.2,1.3) ellipse[x radius=1.85cm,y radius=1.25cm];
\fill[blue!10] (2.2,1.3) ellipse[x radius=1.05cm,y radius=0.68cm];
\draw[blue!60!black,thick] (2.2,1.3) ellipse[x radius=1.05cm,y radius=0.68cm];
\node[blue!60!black] at (2.2,1.65) {core};
\node[orange!70!black] at (2.2,2.25) {broad boundary (rim)};
\fill (1.9,1.1) circle[radius=1.8pt] node[anchor=west,inner sep=2.5pt] {holds};
\fill (3.6,1.0) circle[radius=1.8pt] node[anchor=north,inner sep=2.5pt] {undecided};
\fill (4.45,2.85) circle[radius=1.8pt] node[anchor=east,inner sep=2.5pt] {fails};
\node[orange!70!black] at (2.2,-0.45) {rim width set by map accuracy};
\end{tikzpicture}
\end{minipage}
\caption{The three kernel components. (a) \cts{} gives every operation a signature over core-concept types; a statement whose arguments have the wrong types, such as a distance from a set of objects, is rejected before it is evaluated. (b) \mia{} treats every written value as an interval set by its digits (bars) and compares a claim with the computed value only within a tolerance band declared for the task: a claim inside the band agrees, and one outside is refuted. In GS-QA T26/689, for example, the claimed 3.647~km lies inside the band of $\pm\max(5\%,\,0.5~\text{km})$ around the computed 3.45~km. (c) \gre{} tests whether a statement about spatial relations holds, e.g. whether a point lies in a region, given map accuracy constraints: the relation holds in the core, fails far outside, and is left undecided if it falls within the buffer allowed by the accuracy of the map.}
\label{fig:kernel}
\end{figure}
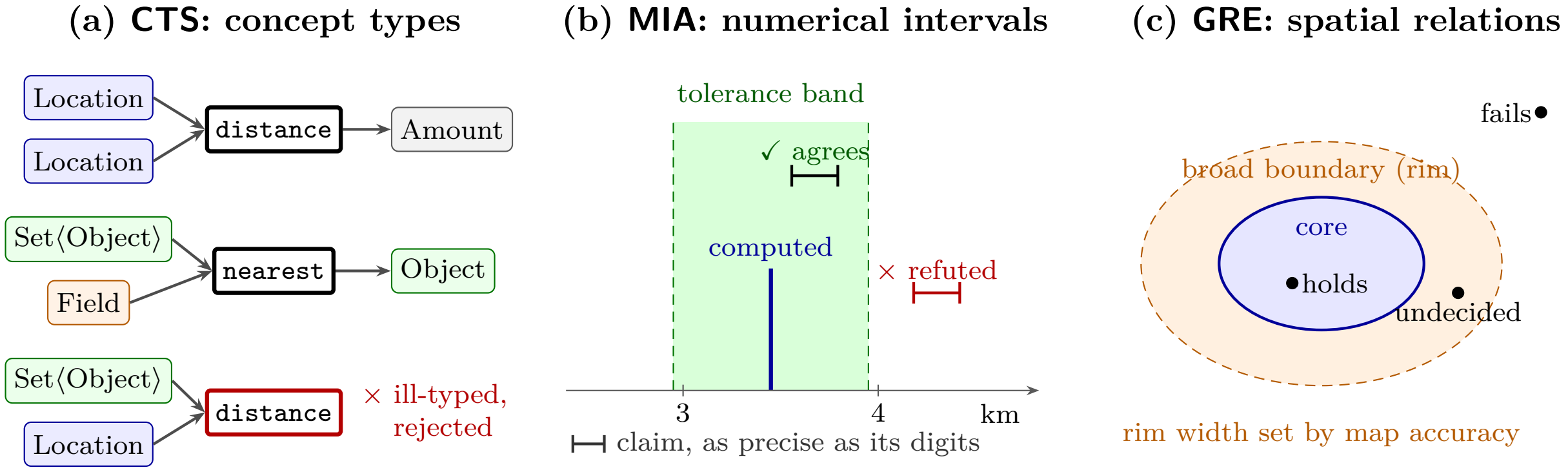

Inside the kernel, three GIScience principles are enforced as invariants by three components (Figure~\ref{fig:kernel}). The type system, \cts{} (Concept Type System), is built on the core concepts of spatial information: every value is a location, object, field, network or event, or an amount derived from them, and every operation has a signature that transforms one concept into another \citep{kuhn2012core,scheider2020ontology}. A distance maps two locations to an amount, and a nearest-place claim maps a set of objects and a distance field to an object; a statement whose arguments have the wrong types is rejected before it is evaluated, as a type checker rejects an ill-typed program. Arithmetic over measurements, \mia{} (Measurement Interval Arithmetic), is interval arithmetic: every value carries the precision interval implied by how it was recorded, so that a coordinate written with five decimals is known to about a meter and one with two decimals to about a kilometer, and a comparison succeeds only within the tolerance declared for the branch. Coordinates and bearings are compared at the precision their written digits state, one unit of the last written digit, so that a claim is never refuted for being rounded as written; a whole number's trailing zeros may themselves be rounding, so ``190 degrees'' is read as stated to the nearest ten. No further tolerance is added. Distances are compared under a \emph{policy}, declared per dataset, that gives a relative bound, an absolute bound, a unit and a source; the larger bound applies (GS-QA declares 5\% with a floor of 0.5~km). The kernel holds no tolerance values of its own: an undeclared policy leaves the comparison undecided rather than guessed. Spatial relations are evaluated by \gre{} (Geometric Relation Evaluator): distances, bearings, containment and nearness are recomputed from the map data; topological relations follow the 9-intersection model conventions \citep{egenhofer1991point}, tested against archived spatial partition data, with an egg-yolk tolerance schema \citep{cohn1996eggyolk}. Containment is checked as \op{holds}(level, region, point, meaning), where the level is a frozen partition (a cities gazetteer, Natural Earth admin-0 including the 12 nautical mile territorial sea, admin-1, continents or land cover) and the meaning is \emph{inside}, \emph{covered by}, or a broad boundary whose rim is set by map accuracy; when the wording does not fix the level, the wrapper emits one interpretation per level. Directions are computed as initial great-circle bearings and tested against the sector each interpretation declares (GS-QA: $\pm22.5^\circ$ around the named direction).

\subsection{Reporter}\label{sec:design-reports}
The \emph{Reporter} organizes the kernel verdicts as a structured natural-language \emph{Outcome}. We design it following five principles:

\begin{itemize}
    \item \textbf{Separating reasoning soundness from conclusion correctness.} A reasoning can contain a false statement but still reach a correct conclusion, while a reasoning with no false statement can still reach a wrong one. \gd{} separates the two clearly: the non-conclusion statements are checked to reach a \emph{reasoning verdict} on reasoning soundness, and the conclusion statement is independently checked to reach a \emph{conclusion verdict} on whether the conclusion can be derived from the reasoning and whether it ties correctly to the goal.
    \item \textbf{Staying conservative on verdicts.} When the kernel cannot reach a verdict, e.g. because the provided information is inadequate or the question is beyond the formal verification capability of the kernel, \gd{} returns an open outcome. It states what is missing and points to a potential solution: supplying more data, or expanding the supported verification goals of \gd{}.
    \item \textbf{Reporting interpretation branches separately.} Following the design of Semantic Wrapper, an outcome always states which interpretation branches it stands on.
    \item \textbf{Tracing back to the first error.} A report names the first invalid statement along the reasoning path to the conclusion statement. Other invalid statements afterwards are listed for reference only. This follows the conventional programming language exception traceback design.
    \item \textbf{Ensuring human-readability.} An outcome is rendered as a typed exception in the style of a Java stack trace, together with a natural-language suggestion on potential solutions, so that a user can easily refer back to the original reasoning, locate the error and fix it.
\end{itemize}

Section~\ref{sec:method-outcomes} implements these principles as a rule table yielding seven Outcomes with a rendered, structured report.

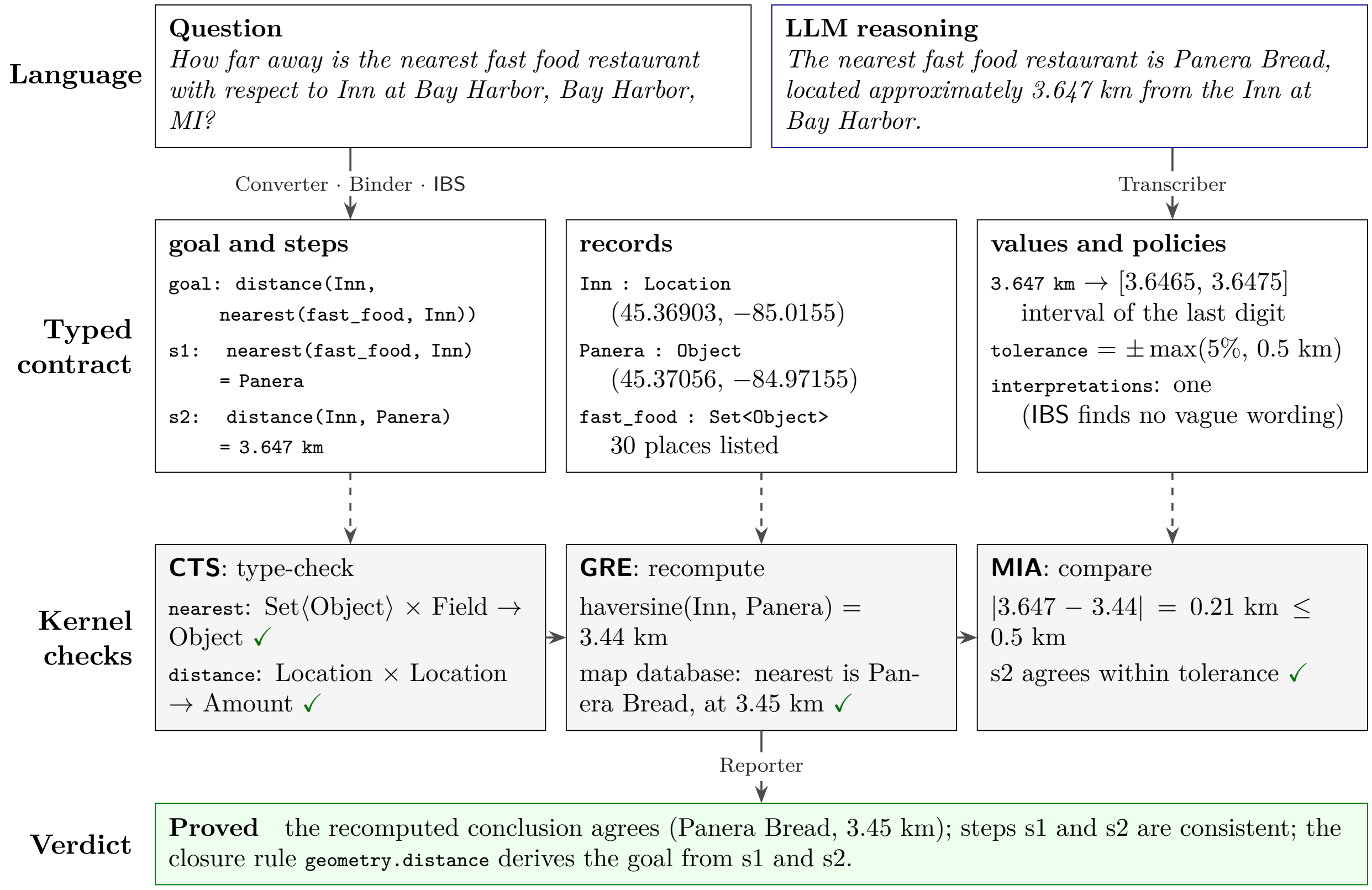
\begin{figure}[!ht]
\centering
\resizebox{\textwidth}{!}{
\begin{tikzpicture}[font=\small,
  band/.style={font=\small\bfseries,align=right,anchor=east,text width=1.7cm},
  bx/.style={align=left,inner sep=5pt,anchor=north,font=\footnotesize,text width=5.0cm},
  flow/.style={-{Stealth[length=2.4mm]},thick,gray!60!black},
  via/.style={font=\scriptsize,fill=white,inner sep=2pt,text=gray!30!black}]
\newcommand{\code}[1]{{\scriptsize\ttfamily #1}}
\newcommand{\ok}{\textcolor{green!45!black}{\checkmark}}
\def\xa{4.7}\def\xb{10.35}\def\xc{16.0}
\node[bx,text width=7.825cm] (q) at (6.1125,0) {\textbf{Question}\\ \emph{How far away is the nearest fast food restaurant with respect to Inn at Bay Harbor, Bay Harbor, MI?}};
\node[bx,text width=7.825cm] (r) at (14.5875,0) {\textbf{LLM reasoning}\\ \emph{The nearest fast food restaurant is Panera Bread, located approximately 3.647~km from the Inn at Bay Harbor.}};
\node[fit=(q)(r),inner sep=0pt] (qrow) {};
\path (qrow.south) ++(0,-1.0) coordinate (row2);
\node[bx] (c1) at (\xa,0 |- row2) {\textbf{goal and steps}\\[3pt]
\code{goal: distance(Inn,}\\\code{\ \ \ \ \ \ nearest(fast\_food, Inn))}\\[2pt]
\code{s1:\ \ \ nearest(fast\_food, Inn)}\\\code{\ \ \ \ \ \ = Panera}\\[2pt]
\code{s2:\ \ \ distance(Inn, Panera)}\\\code{\ \ \ \ \ \ = 3.647 km}};
\node[bx] (c2) at (\xb,0 |- row2) {\textbf{records}\\[3pt]
\code{Inn : Location}\\\hspace*{1.2em}(45.36903, $-$85.0155)\\[2pt]
\code{Panera : Object}\\\hspace*{1.2em}(45.37056, $-$84.97155)\\[2pt]
\code{fast\_food : Set<Object>}\\\hspace*{1.2em}30 places listed};
\node[bx] (c3) at (\xc,0 |- row2) {\textbf{values and policies}\\[3pt]
\code{3.647 km} $\to$ [3.6465, 3.6475]\\\hspace*{1.2em}interval of the last digit\\[2pt]
\code{tolerance} = $\pm\max(5\%,\,0.5~\text{km})$\\[2pt]
\code{interpretations}: one\\\hspace*{1.2em}(\ibs{} finds no vague wording)};
\node[fit=(c1)(c2)(c3),inner sep=0pt] (crow) {};
\path (crow.south) ++(0,-1.0) coordinate (row3);
\node[bx] (k1) at (\xa,0 |- row3) {\textbf{\cts{}}: type-check\\[3pt]
\code{nearest}: Set$\langle$Object$\rangle$ $\times$ Field $\to$ Object \ok\\[2pt]
\code{distance}: Location $\times$ Location $\to$ Amount \ok};
\node[bx] (k2) at (\xb,0 |- row3) {\textbf{\gre{}}: recompute\\[3pt]
haversine(Inn, Panera) = 3.44~km\\[2pt]
map database: nearest is Panera Bread, at 3.45~km \ok};
\node[bx] (k3) at (\xc,0 |- row3) {\textbf{\mia{}}: compare\\[3pt]
$|3.647 - 3.44| = 0.21~\text{km} \le 0.5~\text{km}$\\[2pt]
s2 agrees within tolerance \ok};
\node[fit=(k1)(k2)(k3),inner sep=0pt] (krow) {};
\path (krow.south) ++(0,-1.0) coordinate (row4);
\node[bx,text width=16.3cm] (v) at (\xb,0 |- row4) {\textbf{\outcome{Proved}}\quad the recomputed conclusion agrees (Panera Bread, 3.45~km); steps s1 and s2 are consistent; the closure rule \code{geometry.distance} derives the goal from s1 and s2.};
\newcommand{\rowbox}[4]{\draw[#3,fill=#4] (#1.west |- #2.north) rectangle (#1.east |- #2.south);}
\begin{scope}[on background layer]
\rowbox{q}{qrow}{black}{white}\rowbox{r}{qrow}{blue!60!black}{white}
\rowbox{c1}{crow}{black}{white}\rowbox{c2}{crow}{black}{white}\rowbox{c3}{crow}{black}{white}
\rowbox{k1}{krow}{black}{gray!8}\rowbox{k2}{krow}{black}{gray!8}\rowbox{k3}{krow}{black}{gray!8}
\rowbox{v}{v}{green!45!black}{green!7}
\end{scope}
\node[band] at (1.85,0 |- qrow.center) {Language};
\node[band] at (1.85,0 |- crow.center) {Typed\\contract};
\node[band] at (1.85,0 |- krow.center) {Kernel\\checks};
\node[band] at (1.85,0 |- v.center) {Verdict};
\draw[flow] (\xa,0 |- qrow.south) -- node[via] {Converter $\cdot$ Binder $\cdot$ \ibs{}} (c1.north |- crow.north);
\draw[flow] (\xc,0 |- qrow.south) -- node[via] {Transcriber} (c3.north |- crow.north);
\foreach \a/\b in {c1/k1,c2/k2,c3/k3} \draw[flow,dashed] (\a.south |- crow.south) -- (\b.north |- krow.north);
\draw[flow] (k1.east |- krow.center) -- (k2.west |- krow.center); \draw[flow] (k2.east |- krow.center) -- (k3.west |- krow.center);
\draw[flow] (k2.south |- krow.south) -- node[via] {Reporter} (v.north);
\end{tikzpicture}}
\caption{One case traced through \gd{} (GS-QA T26/689, simplified to two steps out of original 34 steps for visualization clarity). The wrapper turns the question and the reasoning into typed content; each part of the contract is checked by one kernel component; the reporter summarizes the verdict, here the accepting outcome \outcome{Proved} (Section~\ref{sec:method-kernel}). Dashed arrows show which part of the contract each component uses, as in Figure~\ref{fig:architecture}.}
\label{fig:worked-flow}
\end{figure}

\subsection{A worked example}\label{sec:design-example}
As a summary, we use one GS-QA case (T26/689) to walk the readers through the architecture. The question is \emph{How far away is the nearest fast food restaurant with respect to Inn at Bay Harbor, Bay Harbor, MI?} The map context gives the reference place at (45.36903, $-85.0155$) and lists 30 fast food candidates, among them Panera Bread at (45.37056, $-84.97155$). The GPT-4o reasoning lists the coordinates, concludes that ``The nearest fast food restaurant is Panera Bread, located approximately 3.647 km from the Inn at Bay Harbor'', and ends with the conclusion statement \{name: Panera Bread, distance: 3.647 km\}. Figure~\ref{fig:worked-flow} traces how DATUM gives its verification: after the conversion from natural language to typed content, a sound reasoning path is found by the kernel; since there is only one possible branch to check, a "Proved" verdict is issued.

\section{Implementation}\label{sec:method}

This section describes the implementation details of \gd{}. First, we systematically define the core type system (goals, statements, exceptions and contracts) that enables formal verification of spatial reasoning (Section~\ref{sec:method-types}). Then, we describe the verification graph on which we algorithmically check a reasoning step by step (Section~\ref{sec:method-kernel}). Finally, we introduce the outcome categories that we design to deliver critical reasoning information to users (Section~\ref{sec:method-outcomes}). We use a correct reasoning (GS-QA T26/689) and a wrong reasoning (GS-QA T15/2270) as illustrative examples.

\subsection{Type system}\label{sec:method-types}
The kernel only accepts typed objects. We divide the types into four kinds: (1) a \emph{goal type} defines an abstract template of a class of questions (e.g. distance questions); (2) a \emph{statement type} defines an abstract template of a class of reasoning steps (e.g. 1-nearest neighbor --> closest); (3) an \emph{exception types} defines an abstract template of a class of reasoning mistakes (e.g. equal longitude differences do not mean equal distance differences); (4) a \emph{contract type} defines a JSON template that the wrapper complies with to format its outputs.

\subsubsection{Goal types}\label{sec:method-records}\label{sec:method-fields}
The \bnd{} reads the question and the context, and returns a typed goal formatted as a JSON structure under a strict schema. The schema has 31 fields. Some fields hold the discretion of the \bnd{}, e.g. the \emph{envelope} field. Some fields hold the typed records extracted from the question, together with the actual text it is derived from, i.e. the \emph{quote}. In this way, every part of the goal can be traced back to specific wording of the original question, allowing users to manually double-check. Figure~\ref{fig:binder-json} shows the actual output for T26/689, \emph{How far away is the nearest fast food restaurant with respect to Inn at Bay Harbor, Bay Harbor, MI?} The most important field is \emph{envelope}. The complete JSON schema with all 31 fields can be found in Appendix~\ref{app:goal}.

\begin{figure}[!htb]
\centering
\input{figures/binder-json}
\caption{The \bnd{} output for GS-QA T26/689. The 25 fields it leaves empty are not shown (Appendix~\ref{app:goal} lists all 31). Each field keeps the quote where the field value is derived from (``nearest'' for $k=1$, ``fast food restaurant'' for the OpenStreetMap category \op{amenity=fast\_food}).}
\label{fig:binder-json}
\end{figure}

\paragraph{Envelopes} Most spatial questions can be categorized -- for example, any ``nearest'' questions can be formalized as a ranking problem. We call these typed templates for a whole family of questions \emph{envelopes}. Formally, an envelope $E$ is a triple $(\Sigma_E, \tau_E, \Phi_E)$: a signature $\Sigma_E$ listing the fields of information needed for this specific $E$ (e.g. a ``nearest'' question needs at least a list of coordinates and a distance function) and the core-concept types of each field, much like how a signature of a programming language function defines the variables and variable data types (Section~\ref{sec:design-kernel}); a result type $\tau_E$; and a set $\Phi_E$ of optional condition types by which $E$ may restrict its candidates (e.g. a ``nearest'' question may require the maximum distance not to exceed 10 miles).

The \op{geometry} envelope, for example, asks for a direction, a distance or both, from a reference place to a target:
\begin{align*}
\op{geometry}\colon\ & \langle\, \mathit{relation},\ \mathit{reference}{:}\ \mathrm{Object},\ \mathit{target} \,\rangle\\
& \quad\;\to\; \mathrm{Amount}\ (\text{distance})\ \text{or a compass direction},\\
\mathit{target} \;=\; & \ \mathit{to}{:}\ \mathrm{Object} \quad\text{or}\quad \mathrm{nearest}_k \{\, c \in C \mid \varphi(c)\ \text{for every}\ \varphi \in \Phi \,\},\\
& \quad C{:}\ \mathrm{Set(Object)},
\end{align*}
where \textit{relation} is \op{distance}, \op{direction} or \op{direction\_distance}, the target is either a named place or the $k$-th nearest member of a candidate set $C$ (a context listing or an OpenStreetMap category) that passes the conditions $\Phi$, and $\Phi_{\op{geometry}}$ admits a distance limit, a direction, a region and an attribute value. When the question gives a radius instead of asking for the nearest, any member within it is a valid target. 

\paragraph{Typed Goals} A \emph{typed goal} is an envelope with all its required fields filled, $g = (E, \sigma)$, where $\sigma$ maps each field of $\Sigma_E$ to a value of the field's type together with its quote. Fields outside $\Sigma_E$ stay empty. In this paper, we designed seven envelopes that cover a large amount of spatial questions (Table~\ref{tab:goals}). When new categories of spatial questions appear, we can always expand the pool of typed goals.

As an example, case T26/689 has a typed goal as shown below:
\[
g = \mathrm{distance}\bigl(\text{Inn at Bay Harbor},\ \mathrm{nearest}_1\{\, c \in \op{amenity=fast\_food} \,\}\bigr),
\]
with no conditions ($\Phi = \emptyset$). The kernel checks the reasoning against the goal (Section~\ref{sec:method-kernel}).

\begin{table}[t]
\tbl{The seven envelopes of a typed goal.}
{\renewcommand{\arraystretch}{1.25}\begin{tabularx}{\textwidth}{@{}l>{\raggedright\arraybackslash}X>{\raggedright\arraybackslash}X@{}}
\toprule
Envelope & What it asks & Example \\
\midrule
\op{lookup} & one fact about one described place & the opening hours of a museum \\\addlinespace[4pt]
\op{filter} & the candidates that pass every condition; returned as an option, a count, a name or a location & how many cafés within 2~km are open on Sunday \\\addlinespace[4pt]
\op{rank} & candidates ordered by a measure; the $k$-th, least or most & the third-closest hotel to the station \\\addlinespace[4pt]
\op{geometry} & direction, distance or both, from one place to another or to the candidate the conditions keep & how far the nearest fast food restaurant is (T26/689); the direction of a statue from a station (Figure~\ref{fig:opening}) \\\addlinespace[4pt]
\op{route\_walk} & one listed route's steps: a measure, the next step, a count, or what it passes & which road follows a given turn \\\addlinespace[4pt]
\op{schedule} & a trip of legs and stays under a budget, a deadline or closing times & the latest departure that still reaches the last stop before it closes \\\addlinespace[4pt]
\op{blocked} & the available facts cannot decide it (subjective, fact unavailable, no operation) & the most beautiful park nearby; becomes \outcome{Ill-posed} \\
\bottomrule
\end{tabularx}}
\label{tab:goals}
\end{table}

\subsubsection{Statement types}\label{sec:method-ops}
The \tsb{} turns each claim of a reasoning into a typed statement, i.e. a predefined typed operation (Table~\ref{tab:ops}). Each operation has a signature in the core concept data types of \citet{scheider2020ontology}, which \cts{} enforces, and a rule for verifying the validity of the statement based on typed records. When an invalid statement is found, a corresponding type of exception will be raised (Section~\ref{sec:method-hierarchy}). As with envelopes, if a new type of statement cannot be expressed by the existing vocabulary of operations, we can always expand our operation table accordingly.

\begin{table}[t]
\tbl{The typed operations, their signatures in the core concept data types (CCD) of \citet{scheider2020ontology}, and the exceptions they raise. Loc: location; Obj: object; Reg: region; Qlt: quality; Nom, Ord, Itv, Ratio: values on a nominal, ordinal, interval or ratio scale; Bool: truth value; R$n$: a relation of $n$ columns, so R2(Obj, Qlt) is a set of objects with a quality each, and R3(Obj, Ratio, Obj) a network with a cost on each link.}
{\footnotesize\renewcommand{\arraystretch}{1.25}\begin{tabularx}{\textwidth}{@{}>{\raggedright\arraybackslash}p{2.5cm}>{\raggedright\arraybackslash}X>{\raggedright\arraybackslash}p{3.3cm}>{\raggedright\arraybackslash\hyphenpenalty=10000\exhyphenpenalty=10000}p{4.4cm}@{}}
\toprule
Operation & Checks the claim against & CCD signature & Exception \\
\midrule
\op{coordinates\_of} & a place's listed point & Obj $\to$ Loc & CoordinateException \\\addlinespace[4pt]
\op{attribute\_of}, \op{hours} & a listed attribute or opening time & R2(Obj, Qlt) $\times$ Obj $\to$ Qlt & ListedValueException, OpeningHoursException \\\addlinespace[4pt]
\op{open\_at} & opening at a day and time & R2(Obj, Qlt) $\times$ Obj $\times$ Itv $\to$ Bool & OpeningHoursException \\\addlinespace[4pt]
\op{travel}, \op{route\_step} & a leg or route measure by mode & R3(Obj, Ratio, Obj) $\times$ Obj $\times$ Obj $\to$ Ratio & TravelTimeException, TravelDistanceException \\\addlinespace[4pt]
\op{distance\_km} & great-circle distance & Loc $\times$ Loc $\to$ Ratio & DistanceException \\\addlinespace[4pt]
\op{bearing\_deg} & initial bearing & Loc $\times$ Loc $\to$ Itv & BearingException \\\addlinespace[4pt]
\op{in\_direction} & a compass relation, per interpretation & Loc $\times$ Loc $\times$ Nom $\to$ Bool & DirectionException \\\addlinespace[4pt]
\op{within\_km} & inside a radius & Loc $\times$ Loc $\times$ Ratio $\to$ Bool & RadiusException \\\addlinespace[4pt]
\op{in\_region} & region membership & Loc $\times$ Reg $\to$ Bool & RegionException \\\addlinespace[4pt]
\op{kind\_of} & a map category & R2(Obj, Nom) $\times$ Obj $\times$ Nom $\to$ Bool & CategoryException \\\addlinespace[4pt]
\op{nearest\_of}, \op{none\_nearer} & the nearest member of a set that passes the conditions & R2(Obj, Ratio) $\to$ Obj & NearestException, OmittedCandidateException \\\addlinespace[4pt]
\op{count}, \op{none\_other} & the members that pass the conditions & R2(Obj, Bool) $\to$ Ratio & CountException, CounterexampleException \\\addlinespace[4pt]
\op{extreme\_of} & the highest or lowest attribute & R2(Obj, Ord) $\to$ Obj & ExtremeException \\\addlinespace[4pt]
\op{arithmetic}, \op{clock} & expressions over stated numbers & Ratio $\to$ Ratio; Itv $\to$ Itv & CalculationException, ClockException \\
\bottomrule
\end{tabularx}}
\label{tab:ops}
\end{table}

\subsubsection{Exception types}\label{sec:method-hierarchy}
\begin{figure}[!htb]
\centering
\resizebox{\textwidth}{!}{
\begin{tikzpicture}[font=\fontsize{8}{9.6}\selectfont,
  base/.style={draw,rounded corners=1.5pt,align=left,inner sep=3pt,anchor=west},
  root/.style={base,line width=1pt,fill=gray!15,text width=3.85cm},
  kind/.style={base,line width=0.8pt,fill=gray!8,text width=3.45cm},
  fam/.style={base,fill=blue!5,draw=blue!50!black,text width=3.25cm},
  err/.style={base,draw=red!60!black,fill=red!4,text width=4.45cm,inner sep=2pt,font=\fontsize{7.5}{8.5}\selectfont},
  edge/.style={gray!55!black,thick},
  gen/.style={gray!55!black,thick,{Triangle[open,length=2.6mm,width=2.4mm]}-}]
\node[err] (DistanceException) at (12.950,-0.000) {\texttt{DistanceException}};
\node[err] (BearingException) at (12.950,-0.460) {\texttt{BearingException}};
\node[err] (CoordinateException) at (12.950,-0.920) {\texttt{CoordinateException}};
\node[fam] (MeasurementException) at (8.800,-0.460) {\textbf{\texttt{MeasurementException}}\\[-1pt]{\fontsize{6.5}{7.5}\selectfont\itshape a number}};
\draw[gen] (MeasurementException.east) -- ++(0.32,0) coordinate (bMeasurementError);
\draw[edge] (DistanceException.west) -- (DistanceException.west -| bMeasurementError);
\draw[edge] (BearingException.west) -- (BearingException.west -| bMeasurementError);
\draw[edge] (CoordinateException.west) -- (CoordinateException.west -| bMeasurementError);
\draw[edge] (DistanceException.west -| bMeasurementError) -- (CoordinateException.west -| bMeasurementError);
\node[err] (DirectionException) at (12.950,-1.680) {\texttt{DirectionException}};
\node[err] (RadiusException) at (12.950,-2.140) {\texttt{RadiusException}};
\node[err] (RegionException) at (12.950,-2.600) {\texttt{RegionException}};
\node[err] (CategoryException) at (12.950,-3.060) {\texttt{CategoryException}};
\node[fam] (RelationException) at (8.800,-2.370) {\textbf{\texttt{RelationException}}\\[-1pt]{\fontsize{6.5}{7.5}\selectfont\itshape a yes/no relation}};
\draw[gen] (RelationException.east) -- ++(0.32,0) coordinate (bRelationError);
\draw[edge] (DirectionException.west) -- (DirectionException.west -| bRelationError);
\draw[edge] (RadiusException.west) -- (RadiusException.west -| bRelationError);
\draw[edge] (RegionException.west) -- (RegionException.west -| bRelationError);
\draw[edge] (CategoryException.west) -- (CategoryException.west -| bRelationError);
\draw[edge] (DirectionException.west -| bRelationError) -- (CategoryException.west -| bRelationError);
\node[err] (NearestException) at (12.950,-3.820) {\texttt{NearestException}};
\node[err] (OmittedCandidateException) at (12.950,-4.280) {\texttt{OmittedCandidateException}};
\node[err] (CountException) at (12.950,-4.740) {\texttt{CountException}};
\node[err] (CounterexampleException) at (12.950,-5.200) {\texttt{CounterexampleException}};
\node[err] (ExtremeException) at (12.950,-5.660) {\texttt{ExtremeException}};
\node[fam] (SelectionException) at (8.800,-4.740) {\textbf{\texttt{SelectionException}}\\[-1pt]{\fontsize{6.5}{7.5}\selectfont\itshape a claim about a set}};
\draw[gen] (SelectionException.east) -- ++(0.32,0) coordinate (bSelectionError);
\draw[edge] (NearestException.west) -- (NearestException.west -| bSelectionError);
\draw[edge] (OmittedCandidateException.west) -- (OmittedCandidateException.west -| bSelectionError);
\draw[edge] (CountException.west) -- (CountException.west -| bSelectionError);
\draw[edge] (CounterexampleException.west) -- (CounterexampleException.west -| bSelectionError);
\draw[edge] (ExtremeException.west) -- (ExtremeException.west -| bSelectionError);
\draw[edge] (NearestException.west -| bSelectionError) -- (ExtremeException.west -| bSelectionError);
\node[err] (ListedValueException) at (12.950,-6.420) {\texttt{ListedValueException}};
\node[err] (OpeningHoursException) at (12.950,-6.880) {\texttt{OpeningHoursException}};
\node[err] (TravelTimeException) at (12.950,-7.340) {\texttt{TravelTimeException}};
\node[err] (TravelDistanceException) at (12.950,-7.800) {\texttt{TravelDistanceException}};
\node[fam] (ContextReadException) at (8.800,-7.110) {\textbf{\texttt{ContextReadException}}\\[-1pt]{\fontsize{6.5}{7.5}\selectfont\itshape a value read from text}};
\draw[gen] (ContextReadException.east) -- ++(0.32,0) coordinate (bContextReadError);
\draw[edge] (ListedValueException.west) -- (ListedValueException.west -| bContextReadError);
\draw[edge] (OpeningHoursException.west) -- (OpeningHoursException.west -| bContextReadError);
\draw[edge] (TravelTimeException.west) -- (TravelTimeException.west -| bContextReadError);
\draw[edge] (TravelDistanceException.west) -- (TravelDistanceException.west -| bContextReadError);
\draw[edge] (ListedValueException.west -| bContextReadError) -- (TravelDistanceException.west -| bContextReadError);
\node[err] (ClockException) at (12.950,-8.560) {\texttt{ClockException}};
\node[fam] (CalculationException) at (8.800,-8.560) {\textbf{\texttt{CalculationException}}\\[-1pt]{\fontsize{6.5}{7.5}\selectfont\itshape own arithmetic}};
\draw[gen] (CalculationException.east) -- ++(0.32,0) coordinate (bCalculationError);
\draw[edge] (ClockException.west) -- (ClockException.west -| bCalculationError);
\draw[edge] (ClockException.west -| bCalculationError) -- (ClockException.west -| bCalculationError);
\node[kind] (StepException) at (4.650,-4.648) {\textbf{\texttt{StepException}}\\[-1pt]{\fontsize{6.5}{7.5}\selectfont\itshape a statement is false}};
\draw[gen] (StepException.east) -- ++(0.32,0) coordinate (bStepError);
\draw[edge] (MeasurementException.west) -- (MeasurementException.west -| bStepError);
\draw[edge] (RelationException.west) -- (RelationException.west -| bStepError);
\draw[edge] (SelectionException.west) -- (SelectionException.west -| bStepError);
\draw[edge] (ContextReadException.west) -- (ContextReadException.west -| bStepError);
\draw[edge] (CalculationException.west) -- (CalculationException.west -| bStepError);
\draw[edge] (MeasurementException.west -| bStepError) -- (CalculationException.west -| bStepError);
\node[err] (ConclusionMismatchException) at (12.950,-9.320) {\texttt{ConclusionMismatchException}};
\node[err] (InterpretedConclusionException) at (12.950,-9.780) {\texttt{InterpretedConclusionException}};
\node[err] (UncheckedConclusionException) at (12.950,-10.240) {\texttt{UncheckedConclusionException}};
\node[kind] (ConclusionException) at (4.650,-9.780) {\textbf{\texttt{ConclusionException}}\\[-1pt]{\fontsize{6.5}{7.5}\selectfont\itshape no single statement to blame}};
\draw[gen] (ConclusionException.east) -- ++(0.32,0) coordinate (bConclusionException);
\draw[edge] (ConclusionMismatchException.west) -- (ConclusionMismatchException.west -| bConclusionException);
\draw[edge] (InterpretedConclusionException.west) -- (InterpretedConclusionException.west -| bConclusionException);
\draw[edge] (UncheckedConclusionException.west) -- (UncheckedConclusionException.west -| bConclusionException);
\draw[edge] (ConclusionMismatchException.west -| bConclusionException) -- (UncheckedConclusionException.west -| bConclusionException);
\node[root] (SpatialReasoningException) at (0.000,-7.214) {\texttt{SpatialReasoningException}};
\draw[gen] (SpatialReasoningException.east) -- ++(0.32,0) coordinate (bSpatialReasoningException);
\draw[edge] (StepException.west) -- (StepException.west -| bSpatialReasoningException);
\draw[edge] (ConclusionException.west) -- (ConclusionException.west -| bSpatialReasoningException);
\draw[edge] (StepException.west -| bSpatialReasoningException) -- (ConclusionException.west -| bSpatialReasoningException);
\node[anchor=north west,align=left,font=\fontsize{7}{8.4}\selectfont] at (0.000,-11.050) {%
\tikz[baseline=-3pt]\draw[gen](0,0)--(0.6,0); inherits from (UML generalization)\qquad
\tikz[baseline=-3pt]\node[err,minimum width=4mm,text width=,minimum height=2.6mm]{}; concrete exception, raised by a check};
\end{tikzpicture}}
\caption{The exception hierarchy of \gd{}, drawn as a UML class hierarchy: each class inherits from the class its arrow points to. The 20 concrete classes on the right are the ones raised; the classes to their left are abstract exception families.}
\label{fig:exceptions}
\end{figure}
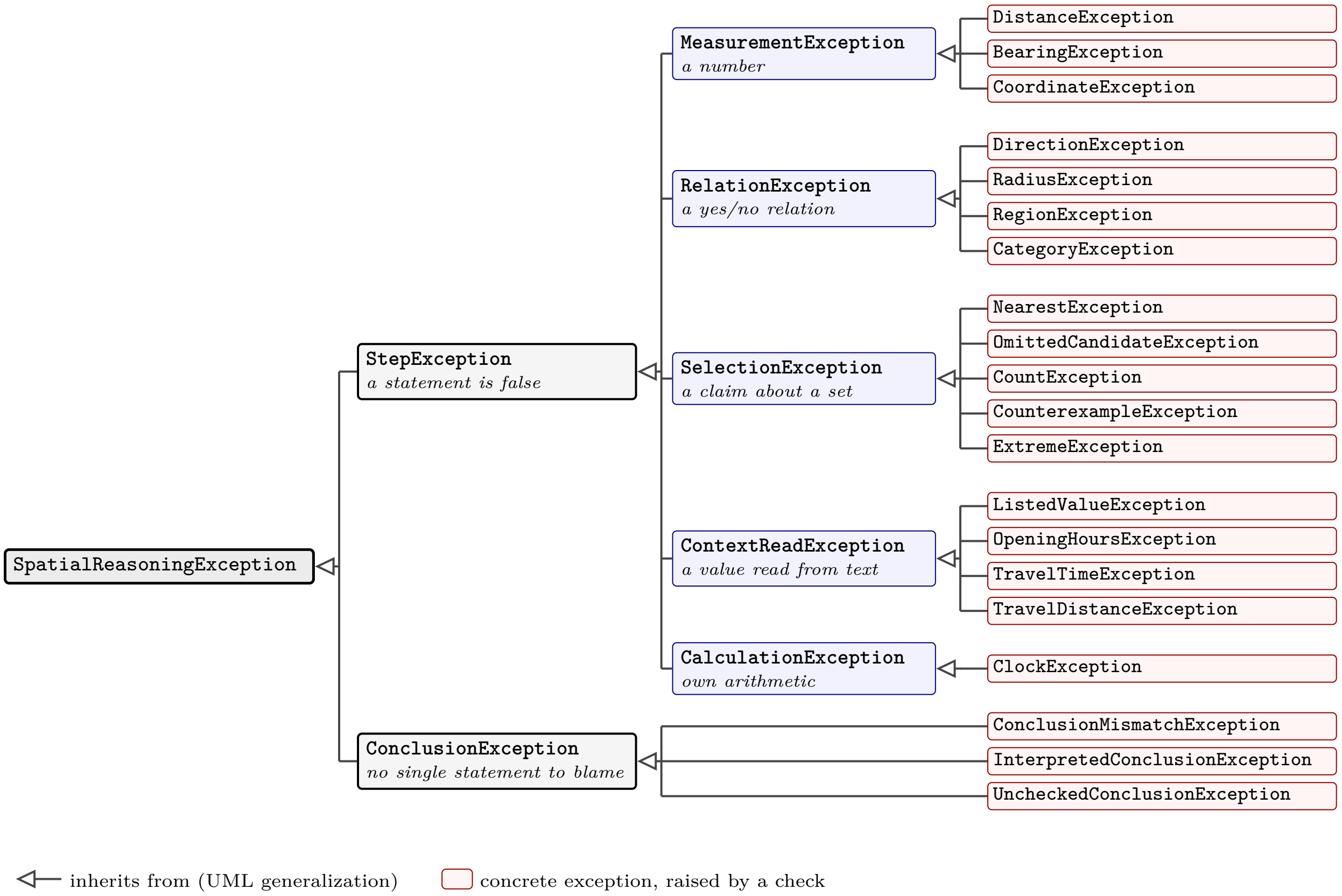

Every check that fails raises a typed exception, and, as in Java, the exceptions form a class hierarchy (Figure~\ref{fig:exceptions}): 20 concrete classes under seven abstract families, all rooted in \texttt{SpatialReasoningException}. The two direct subclasses of the root separate what went wrong. A \texttt{StepException} means a statement is false. A \texttt{ConclusionException} means the conclusion fails, or cannot be checked, while no reasoning statement is to blame. \texttt{StepException} is split again by how the step is identified as false: numerical inconsistency is classified as \texttt{MeasurementException}; false spatial relations are classified as \texttt{RelationException}; a false choice of candidates is classified as \texttt{SelectionException}; the step using wrong context records is classified as \texttt{ContextReadException}; and the reasoning's own arithmetic fault is classified as \texttt{CalculationException}. Each operation raises one concrete type of exceptions (Table~\ref{tab:ops}). Exceptions are further marked as \emph{errors} if the exception directly leads to a failed conclusion, or \emph{warnings} if otherwise.

\subsubsection{Typed contracts}\label{sec:method-contracts}
The contracts designed in Section~\ref{sec:design-contract} are implemented as five concrete rules listed in Table~\ref{tab:contracts}. Every time the wrapper outputs typed content, the five checks will be run independently on it and decide collectively whether to accept this input. Major issues will skip the kernel verification: for example, a binding that fails to identify a valid typed goal will yield an \outcome{Open} outcome, marking \bnd{} as the root cause (Section~\ref{sec:method-outcomes}). Minor binding issues, such as \tsb{} does not find in-text precision information about a stated value, which is very common in spatial QA datasets, will mark this quote as \emph{uncheckable} but still pass the content to the kernel. What typed contracts include depends solely on what the kernel accepts -- if new goals, new operations or new spatial databases are added to the kernel, corresponding contracts need to be implemented accordingly.

\begin{table}[!htbp]
\tbl{The typed contracts implemented in \gd{}.}
{\begin{tabularx}{\textwidth}{@{}>{\raggedright\arraybackslash}p{2.6cm}>{\raggedright\arraybackslash}X>{\raggedright\arraybackslash}X@{}}
\toprule
Contract & Passes & Fails \\
\midrule
Schema & The required JSON fields are complete & Wrong JSON structure, undefined fields, unacceptable value formats \\\addlinespace[4pt]
Quote & All quotes are spans of exact wording in the original reasoning. & Paraphrases or hallucinated quotes. \\\addlinespace[4pt]
Number & Numerical values with stated precision that match the original text. & Inconsistent numerical values or missing precision information. \\\addlinespace[4pt]
Operation & Operations acceptable by the kernel with correct type declarations. & A wrong or undefined operation. \\\addlinespace[4pt]
Name & Place names that match a record in the kernel's database. The same thing as geocoding. & An undefined or ambiguous place name. \\
\bottomrule
\end{tabularx}}
\label{tab:contracts}
\end{table}

\subsection{Verification graph}\label{sec:method-kernel}
The technique that makes formal verification of a spatial reasoning possible is a \emph{verification graph}: a directed acyclic graph whose nodes are the transcribed statements and whose edges are the dependence chains between statements. Context records, such as coordinates and place names, enter the graph as node arguments. Each node must be one typed operation from Table~\ref{tab:ops}. 

\begin{figure}[!htb]
\centering
\includegraphics[width=\textwidth]{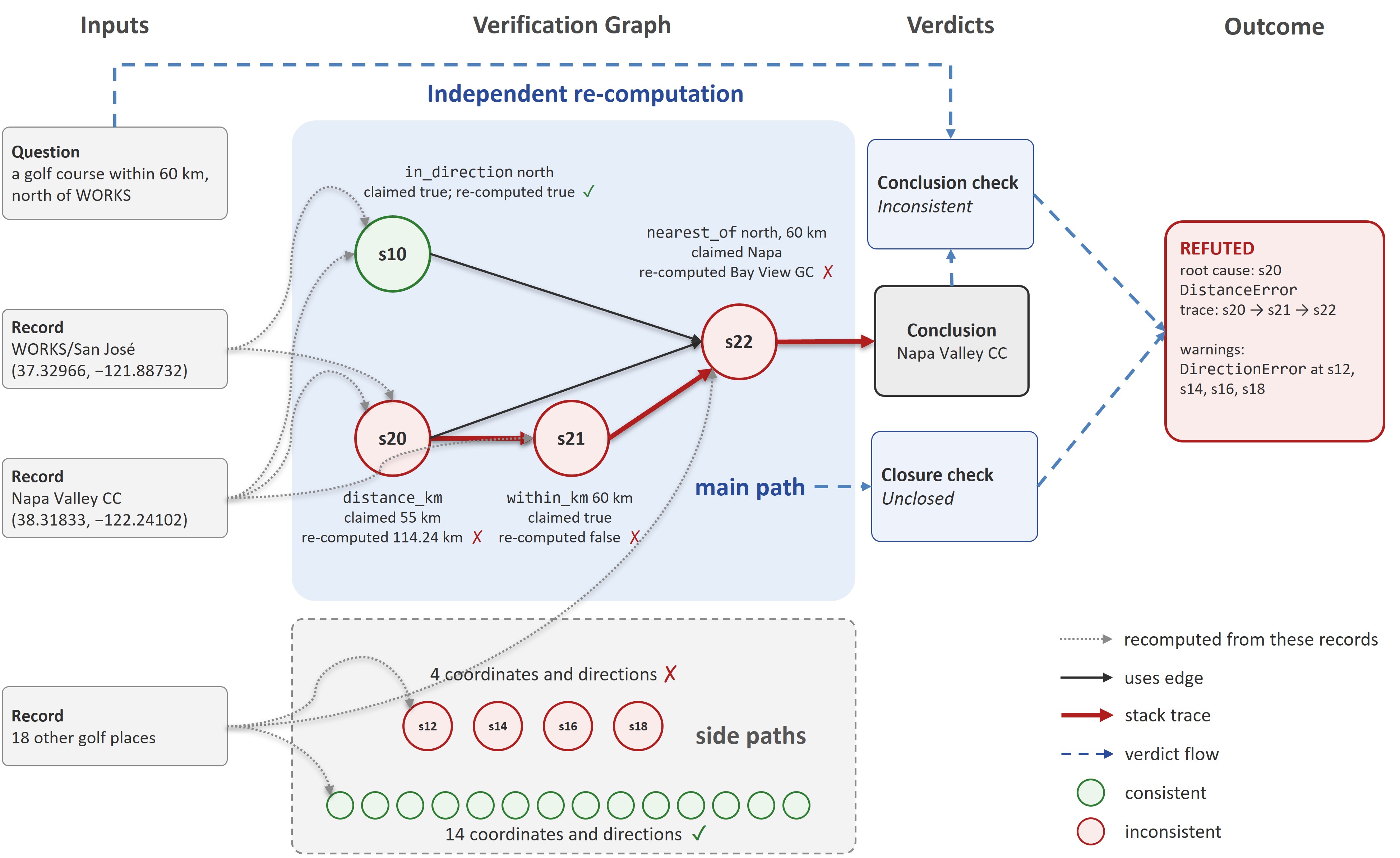}
\caption{The verification graph of GS-QA T15/2270. Dotted gray edges show the context records each statement is re-computed from. Solid black edges are the \emph{uses} edges of the transcription. The main path holds every statement the conclusion depends on, found by following the \emph{uses} edges back from the conclusion statement. Other statements lie on side paths. The closure check and the conclusion check yield two verdicts, which are combined as an outcome (Section~\ref{sec:method-outcomes}). The first inconsistent statement on the main path is the root cause, \texttt{s20}. Inconsistent statements on side paths are reported as warnings.}
\label{fig:check-graph}
\end{figure}

The kernel checks every node in two ways. First, the arguments a node consumes must have the correct types its operation requires (e.g. a distance operation needs two locations, not a road network). An ill-typed node is rejected immediately. Second, the claimed values (e.g. a statement may say ``the distance between A and B is 3.5km'') are compared with the context records for any inconsistency. Notice that we \emph{never} use the claimed values from other statements, so that the error in one node will not propagate to others. Then, the kernel traverses the graph to find a path of valid statements that flow from inputs to the conclusion, which guarantees a valid reasoning; otherwise, the reasoning fails. Figure~\ref{fig:check-graph} plots the verification graph of GS-QA T15/2270 as a concrete example. The rest of this subsection will describe how such a graph is constructed and checked step by step.

\paragraph{Construction} The graph is built from transcribed statements. The \tsb{} returns a list of typed statements, each with an identifier, an operation from Table~\ref{tab:ops}, arguments (place names, coordinates, directions, radius, regions, etc.), values the statements claims, the exact quotes to the original reasoning text, and a \emph{uses} list naming the earlier statements it relies on. Each statement becomes a node, and each entry of its \emph{uses} list becomes an edge. The kernel may add \emph{implicit} nodes if the original reasoning text skips important steps.

\paragraph{Statement check} Every node is first checked by its operation signature, just like the typed check of a programming language function call. Table~\ref{tab:ops} lists the operations with their core-concept signatures (\cts{}) and the exceptions they raise. Then, the values each node claims will be compared against the ones computed directly from the arguments and the operation. We never use claimed values from other statements, so that we will not confuse the error from the node we are checking with the error inherited from another false statement.

Within an interpretation branch, every statement has a fixed meaning, so a statement check exactly returns one of three verdicts: \emph{Consistent} (the signature is consistent with both its upstream and downstream nodes, and the claimed values agree with the context records), \emph{Inconsistent} (inconsistent signature or disagreeing claimed values), or \emph{Uncheckable} (the context does not supply enough information that the operation needs). In Figure~\ref{fig:check-graph}, 15 nodes are consistent and 7 are inconsistent.

\paragraph{Main path and side paths} The \emph{main path} is the set of nodes from which the conclusion statement can be reached along \emph{uses} edges, i.e. the statement chains the conclusion depends on. It is found by tracing back from the conclusion statement to the inputs. Other nodes lie on \emph{side paths}, e.g. a coordinate look-up operation that is never used. We call the consistent statements \emph{lemmas}; they are the statements a closure rule uses. In T15/2270 the main path consists of s10, s20, s21, s22; the other eighteen statements lie on side paths, among which four false direction claims exist but never affect the reasoning validity. The distinction between a main path and side paths matters because it guarantees that a false statement on a side path -- in many cases hallucinated reasoning steps -- will not trigger a false alarm.

\paragraph{Conclusion check} Independently of the verification graph, the kernel uses a pre-defined template verification graph for each goal type and run the statement checks over the template. We call the result \emph{re-computation}. Then we compare the re-computation with the conclusion statement, giving one of three verdicts: \emph{Consistent}, \emph{Inconsistent}, or \emph{Uncheckable} when the re-computation cannot be completed with the information supplied. Notice that the re-computation is only a kernel reference -- being consistent with re-computation does not guarantee a correct conclusion, nor inconsistent guarantee an incorrect conclusion. For T26/689 the re-computation gives Panera Bread at 3.45~km, which agrees with the input; for T15/2270 the re-computation gives Bay View Golf Club, different from Napa Valley Country Club.

\paragraph{Closure check} A reasoning whose statements are all consistent may still not reach a correct conclusion, because it may have misunderstood the question, i.e.\ it does not tie correctly to the goal. We call this verification a \emph{closure check}. Each envelope has its own \emph{closure rules}, listed in Table~\ref{tab:closure}. A rule lists the lemmas that are required to establish the corresponding typed goal. A closure check returns one of three verdicts: \emph{Closed}, \emph{Unclosed}, or \emph{Uncheckable} (meaning that no rule is implemented to cover the goal). If a rule exists for the goal, a reasoning closes if and only if (1) it has all the lemmas the rule requires, and (2) no statement on its main path is \emph{Inconsistent}; otherwise it is unclosed. For T26/689, the rule \op{geometry.distance} closes the goal from the lemmas \op{nearest\_of} and \op{distance\_km}; for T15/2270 no rule can close it because s21 and s22 are inconsistent. Because a closed reasoning must arrive at the conclusion, a conclusion statement value that mismatches the main path calculation (e.g. the main path arrives at ``3.45~km'' but the conclusion says ``3.74~km'', a common hallucination) yields \emph{Unclosed}. Notice that this is different from a conclusion check: the closure check verifies whether the conclusion statement matches its own reasoning steps; the conclusion check verifies whether the conclusion statement matches a third-party conclusion.

\begin{table}[!htbp]
\tbl{The closure rules of the kernel. A rule closes a goal when the consistent statements supply the lemmas it needs, and the value they establish matches the conclusion statement. $\Phi$: the goal's conditions; $r$: the reference place. Rules are grouped by family, one per envelope; a rule is named by its family and its name, e.g.\ \op{rank.nearest}.}
{\footnotesize\renewcommand{\arraystretch}{1.25}\begin{tabularx}{\textwidth}{@{}>{\raggedright\arraybackslash}p{4.1cm}>{\raggedright\arraybackslash}X>{\raggedright\arraybackslash}X@{}}
\toprule
Rule & Goal & Lemmas \\
\midrule
\multicolumn{3}{@{}l}{\emph{lookup}}\\
\op{attribute} & one field of one place (a rating, an address, a yes/no flag) & \op{attribute}(place, field) \\
\op{open} & whether a place is open on a day, at a time & \op{open}(place, day, time) \\\midrule
\multicolumn{3}{@{}l}{\emph{filter}}\\
\op{count} & how many listed candidates pass $\Phi$ & \op{count}($r$, $\Phi$), under exactly $\Phi$ \\
\op{member} & a named candidate that passes $\Phi$ & each condition of $\Phi$ decided true for it, or a nearest or extreme lemma that names it under at least $\Phi$ \\
\op{option} & the offered option that passes $\Phi$ & as \op{filter.member}, and for every other option a condition it fails, or \op{none\_other}($r$, option, $\Phi$) \\\midrule
\multicolumn{3}{@{}l}{\emph{rank}}\\
\op{nearest} & the nearest candidate under $\Phi$ & \op{nearest}($r$, $\Phi$), under exactly $\Phi$ \\
\op{distance\_by\_cases} & the $k$-th nearest offered option & \op{distance}($r$, option) for every option \\
\op{pair\_distance\_by\_cases} & the closest or farthest offered pair & \op{distance}($a$, $b$) for every pair \\
\op{extreme} & the candidate with the highest or lowest attribute & \op{extreme}($r$, field, order, $\Phi$) \\
\op{cardinal} & the option in a named cardinal direction & a direction lemma for every option \\
\op{routes} & the fastest or shortest listed route & the measure of every listed route; no tie \\\midrule
\multicolumn{3}{@{}l}{\emph{geometry}}\\
\op{distance} & the distance between two places, or to the nearest candidate & \op{distance}($a$, $b$), with \op{nearest}($r$, $\Phi$) when $b$ is a candidate \\
\op{direction} & the compass direction & \op{direction}($a$, $b$, word) holding, with \op{nearest} as above \\
\op{bearing} & the direction, as a bearing & \op{bearing}($a$, $b$), with \op{nearest} as above \\
\op{direction\_distance} & both & a direction and a distance lemma \\\midrule
\multicolumn{3}{@{}l}{\emph{route} (envelope \op{route\_walk})}\\
\op{measure} & one listed route's length or travel time & \op{travel}($a$, $b$, km or minutes) \\\midrule
\multicolumn{3}{@{}l}{\emph{trip} (envelope \op{schedule})}\\
\op{travel\_total} & the total travel time of a trip & \op{travel} for every leg; the sum is exact \\
\bottomrule
\end{tabularx}}
\label{tab:closure}
\end{table}

\subsection{Outcomes}\label{sec:method-outcomes}
Following the principles of Section~\ref{sec:design-reports}, \gd{} returns one of seven outcomes, which separate whether the reasoning is sound (the statement and closure checks) from whether the conclusion is correct (the conclusion check). For each interpretation branch, \outcome{Proved} and \outcome{Confirmed} accept the conclusion statement; \outcome{Refuted} and \outcome{Falsified} reject it; \outcome{Open} and \outcome{Ill-posed} cannot decide, for lack of information or an undefined goal. For the cases where there are different interpretations, \outcome{Contingent} accepts or rejects conditionally.

\paragraph{Within an interpretation branch} Three results decide an outcome: the closure verdict, the conclusion verdict, and whether a statement on the main path is \emph{Inconsistent} (a \emph{main-path inconsistency}). Table~\ref{tab:outcomes} lists every combination. Reading down a column, a \emph{Consistent} conclusion is accepted: as \outcome{Proved} when the reasoning closes and as \outcome{Confirmed} when it does not close (e.g. an incomplete reasoning). An \emph{Inconsistent} conclusion is rejected: as \outcome{Refuted} when the error is located on the main path (e.g. a distance the conclusion depends on computed as a Euclidean distance on large latitude and longitude differences), and as \outcome{Falsified} when no error is located on the main path (e.g. the reasoning text skips the steps where it goes wrong). An \emph{Uncheckable} conclusion is still accepted by a closed reasoning (\outcome{Proved}) or rejected by a main-path inconsistency (\outcome{Refuted}); otherwise the branch is \outcome{Open}, or \outcome{Ill-posed} when the cause lies in the undefined goal itself. 

The \emph{Closed} but \emph{Inconsistent} cell is a special case: that means the input reasoning closes but it does not match the kernel re-computation. As we discussed, such mismatch does not flag a fault neither in the input reasoning nor in the kernel re-computation; the only way to settle the dispute is to refer to human judgment. Thus we leave it as \outcome{Open}. In our experiments, such case never occurred.

\begin{table}[!htb]
\tbl{Outcome of one interpretation branch, by conclusion verdict (rows), closure verdict, and main-path inconsistency (columns). Every combination has exactly one outcome.}
{\renewcommand{\arraystretch}{1.3}\begin{tabularx}{\textwidth}{@{}l>{\centering\arraybackslash}X>{\centering\arraybackslash}X>{\centering\arraybackslash}X@{}}
\toprule
 & \multicolumn{3}{c}{Closure verdict} \\ \cmidrule(l){2-4}
 & \emph{Closed} & \multicolumn{2}{c}{\emph{Unclosed} or \emph{Uncheckable}} \\ \cmidrule(l){3-4}
Conclusion verdict & & main-path inconsistency & no main-path inconsistency \\
\midrule
\emph{Consistent} & \outcome{Proved} & \outcome{Confirmed} & \outcome{Confirmed} \\
\emph{Inconsistent} & \outcome{Open (special)} & \outcome{Refuted} & \outcome{Falsified} \\
\emph{Uncheckable} & \outcome{Proved} & \outcome{Refuted} & \outcome{Open}, or \outcome{Ill-posed} if the goal is undefined \\
\bottomrule
\end{tabularx}}
\label{tab:outcomes}
\end{table}

\paragraph{Across interpretation branches} When the question admits several interpretations, the per-branch outcomes are combined following supervaluation semantics. An outcome that every branch agrees is the outcome of the case, and a conclusion that some interpretations accept and other rejects is \outcome{Contingent}; the report then lists the outcome of every branch.

\paragraph{Warning Tags} When the side paths contain \emph{Inconsistent} statements, we list them as warnings appended to the exception traceback, just like in Java.

\paragraph{Reports}\label{sec:method-exceptions}
The reporter renders the graph's verdicts as a Java-style stack trace: the exception raised at the root cause, one \texttt{at} frame for each node of the trace down to the conclusion, the re-computation, the warnings, and a potential fix suggestion based on the exception class (Section~\ref{sec:method-hierarchy}). Every line is deterministically rendered from typed states of the kernel, not generated by LLMs. An example report reads:

\begin{lstlisting}
gsqa/T13/2013: low-risk warning [GD-L201]: a step is wrong, but the conclusion still agrees
NearestException: William Paterson University Galleries is not the nearest place to Russ Berrie
  Institute for Professional Sales, Wayne, NJ: silentsiren (277 deg, 3.40 km) is nearer;
  the reasoning names it but misjudges it
    at s22 nearest_of(from=Russ Berrie Institute ...)  "The closest gallery is the William
       Paterson University Galleries"
    at s25 nearest_of(from=..., radius_km=80)  "Therefore, the answer is the William
       Paterson University Galleries."
    at conclusion {"name": "William Paterson University Galleries", ...}
Fix: compare the distances in s22 again; the nearer place named above passes the same
  conditions
\end{lstlisting}

\section{Experiments}\label{sec:setup}\label{sec:results}
A verifier earns trust in an autonomous pipeline only if three properties hold: it accepts right reasoning and rejects wrong reasoning, its report helps fix the errors it finds, and its reliability does not hinge on the LLM it uses. The experiments test whether \gd{} does better than existing baselines on each of these properties. (Q1) Does \gd{} tell wrong conclusions from right ones more reliably than the baselines (Section~\ref{sec:res-main})? (Q2) Do \gd{}’s reports locate errors accurately, and are they useful enough for an LLM to repair a wrong reasoning (Sections~\ref{sec:res-injection} and~\ref{sec:res-repair})? (Q3) Do these results depend on the LLMs we use (Section~\ref{sec:res-ablation})? We also compare what each method costs in dollars, counting not only its API expense but also the expected cost of the wrong conclusions it accepts. \gd{}’s API expense is higher than a typical LLM-as-judge’s, yet it has the lowest expected cost of all verifiers on every model, because the wrong conclusions the other methods accept cost far more than extra verification (Section~\ref{sec:res-main}).

\subsection{Experiment setup}\label{sec:exp-setup}
We evaluate \gd{} and the baselines on 585 Spatial CoT reasonings from two spatial question-answering benchmarks, with six LLMs from three providers in two price tiers. This subsection describes the data (Section~\ref{sec:data}), the baselines (Section~\ref{sec:baselines}) and the metrics (Section~\ref{sec:metrics}).

\subsubsection{Datasets and Spatial CoT reasoning}\label{sec:data}
We evaluate \gd{} on two spatial question answering datasets, \emph{MapEval} \citep{dihan2024mapeval} and \emph{GS-QA} \citep{saeedan2026gsqa}. Together they cover two major types of spatial reasoning tasks: multi-choice questions (e.g. nearest place, trip planning) and computation questions (e.g. travel distance, spatial relations). MapEval tests reasoning over curated, heterogeneous context with closed-end options, while GS-QA tests reasoning over raw map data with open-end value calculation.

\begin{itemize}
\item \textbf{MapEval} \citep{dihan2024mapeval} evaluates the geospatial reasoning of foundation models with 700 multiple-choice questions over 180 cities in 54 countries, in textual, API-based and visual tracks. Its authors found that even the strongest models trail people by more than 20 percentage points, and fail most often on distances, directions, routes and place-specific facts. We use its 300 textual questions, which gives spatial context in plain language (places, addresses, ratings, opening hours, travel times and distances) and asks about nearby places (83), place information (64), routing (66) and trip planning (67), plus 20 questions that the context cannot answer.
\item \textbf{GS-QA} \citep{saeedan2026gsqa} has 2,800 questions from 28 templates over OpenStreetMap and Wikipedia data for the United States, with spatial questions involving place names (1,200), locations (800), directions, counts and distances (200 each), and areas and lengths (100 each). Benchmarks show that LLMs do well on simple predicates with name answers but drop on complex spatial relations and numeric answers. We use a random subset of 300 questions, covering 25 of its 28 templates and five of its seven answer types (names 102, locations 67, directions 52, distances 48, counts 16). Since the GS-QA dataset does not supply the spatial context needed for question answering (e.g. place names, coordinates), we build one for each question from an OpenStreetMap database. 285 questions remain usable after the context construction.
\end{itemize}

The reasonings we use for evaluation come from Spatial CoT \citep{liu2026spatialcot}, the state-of-the-art in LLM spatial question answering. On MapEval's textual questions it beats chain of thought and self-consistency with each closed-source model it tests, reaching 91.0\% accuracy with GPT-5 and 79.3\% with GPT-4o, against 88.3\% and 75.3\% for chain of thought. Since this paper is concerned with locating spatial reasoning flaws, we chose their GPT-4o benchmark and applied it to GS-QA, which generate a significant amount of errors for us to test with.
Table~\ref{tab:data} lists the statistics of the reasoning benchmark.

\begin{table}
\tbl{Reasoning benchmark. A reasoning is \emph{wrong} if its conclusion statement disagrees with the published key under the dataset's declared tolerance policy.}
{\begin{tabular}{lrrr}
\toprule
Dataset & \# Reasoning & \# Wrong & GPT-4o Accuracy \\
\midrule
MapEval & 300 & 61 & 79.7\% \\
GS-QA & 285 & 97 & 66.0\% \\
\midrule
All & 585 & 158 & 73.0\% \\
\bottomrule
\end{tabular}}
\label{tab:data}
\end{table}

\subsubsection{Baselines}\label{sec:baselines}
We compare \gd{} with existing methods that check whether an LLM's reasoning can be trusted. Solving the task again is the most direct independent check, since a model asked to review its own reasoning without outside feedback rarely finds its errors \citep{huang2024selfcorrect}. The dominant way is to ask an LLM to judge another model's output \citep{zheng2023judging,gu2024survey}, i.e.\ \emph{LLM-as-judge}. Judging step by step moves forward to locating errors in a reasoning rather than only judging the conclusion \citep{lightman2024verify,zheng2025processbench}. Repeating or combining methods, such as by sampling several solutions \citep{wang2023selfconsistency}, asking a panel of judges from different models \citep{verga2024juries}, or checking a derived conclusion against the reasoning that produced it \citep{weng2023selfverification}, is the standard way to make judgment more robust. We follow these conventions and adopt baselines in three groups.

\paragraph{Re-solve} The model solves the task again without seeing the original reasoning, and the original conclusion is accepted if it matches the new conclusion. The model returns only its conclusion, without working.

\paragraph{LLM-as-judge} The model reads the task, the original reasoning and the conclusion. It returns \emph{accept}, \emph{reject} or \emph{cannot decide}. We evaluate three variants:
\begin{itemize}
\item \textbf{Generic judge}: A plain LLM-as-judge prompt \citep{zheng2023judging} that asks only whether the conclusion is correct, with a short explanation.
\item \textbf{DATUM-informed judge} (\emph{informed judge} in short): A prompt that explicitly asks the LLM to locate errors that match the exception types of \gd{}. It is a stronger baseline than plain LLM-as-judge.
\item \textbf{Step-level judge}: The reasoning is split into numbered sentences by a fixed rule, and the LLM judge labels each one correct, incorrect or unverifiable before ruling on the conclusion, in the manner of process verification \citep{lightman2024verify,zheng2025processbench}. It is a stronger baseline than plain LLM-as-judge.
\end{itemize}

\paragraph{Combined runs} Several judgments decide together.
\begin{itemize}
\item \textbf{Re-solve + generic judge}, \textbf{Re-solve + informed judge} and \textbf{Re-solve + step-level judge}: A re-solve run and an LLM-as-judge run are conducted independently. The verdict is \emph{accept}/\emph{reject} if the two verdicts agree, and \emph{cannot decide} otherwise. This combination mimics the design of DATUM: the check of conclusion (Re-solve) is independent of the check of reasoning (LLM-as-judge). We call these three the \emph{re-solve + judge} baselines.
\item \textbf{Jury of judges} \citep{verga2024juries}: A jury consisting of one of the three LLM-as-judge variants -- generic, informed and step-level -- conducted on models from different providers (See Table \ref{tab:models}). The majority of the three decides.
\item \textbf{Self-consistency re-solve} \citep{wang2023selfconsistency}: Five independent re-solve runs decide together, either by majority or by unanimity. In this paper we only conducted self-consistency re-solve on GPT-6 Sol and GPT-6 Luna as an ablation study, because it costs too much yet improves very little over a simple single re-solve run (Section~\ref{sec:res-ablation}, Table~\ref{tab:repeated}).
\end{itemize}

\paragraph{Models} Every method runs on six models from three major LLM providers (Table~\ref{tab:models}): the large and the small general-purpose model each provider offers at the time of writing, grouped into a large and a small tier. Every model runs by default with its hidden reasoning off. Turning it on is an ablation on GPT-6 Sol (Section~\ref{sec:res-ablation}). The prompts are given in Appendix~\ref{app:prompts}.

\begin{table}
\tbl{Models grouped by API prices in US dollars per million input and output tokens.}
{\begin{tabular}{llll}
\toprule
Tier & Provider & Model & Price (input / output) \\
\midrule
Large & OpenAI & GPT-6 Sol & 2.00 / 10.00 \\
 & Anthropic & Claude Sonnet 5.5 & 2.00 / 10.00 \\
 & Google & Gemini 4 Argon & 2.00 / 10.00 \\
\midrule
Small & OpenAI & GPT-6 Luna & 0.10 / 0.50 \\
 & Anthropic & Claude Haiku 5.5 & 0.10 / 0.50 \\
 & Google & Gemini 3.1 Flash-Lite & 0.25 / 1.50 \\
\bottomrule
\end{tabular}}
\label{tab:models}
\end{table}

\subsubsection{Metrics}\label{sec:metrics}
All methods make one of three decisions on a reasoning: \emph{accept}, \emph{reject}, or \emph{cannot decide}. For \gd{}, \emph{accept} means the outcome is \outcome{Proved} or \outcome{Confirmed}; \emph{reject} means \outcome{Refuted} or \outcome{Falsified}; \emph{cannot decide} means \outcome{Open}, \outcome{Ill-posed} or \outcome{Contingent}; a baseline cannot decide when it says so or returns an invalid response.

Among the reasonings that a verifier can decide, we define \emph{positive} as a \emph{reject} decision, i.e. locating an error, and \emph{negative} as a \emph{accept} decision. Therefore, a \emph{false positive} is where the verifier mistakes a correct reasoning for wrong, and a \emph{false negative} is where a wrong reasoning for correct. For autonomous GIS, the false negatives are the most destructive -- it silently lets errors propagate through the pipelines. The false positives mostly cause cost problems -- reviewing correct reasoning wastes human and machine resources.

Table~\ref{tab:confusion} defines the quantities we need: $N$ is the total number of reasonings; $n_{\mathrm{TP}}$, $n_{\mathrm{TN}}$, $n_{\mathrm{FP}}$ and $n_{\mathrm{FN}}$ are the numbers of true positives, true negatives, false positives and false negatives, respectively; $n_{\mathrm{U}}$ is the number of reasonings that a verifier cannot decide. Table~\ref{tab:metrics} defines the measures we compute from these counts. Together they test \emph{verification reliability}: how much a method decides, how often its decisions are right, and how much risk it leaves.

\begin{table}
\tbl{Definitions of verification quantities.}
{\begin{tabular}{l ccc}
\toprule
 & Reject & Accept & Cannot decide \\
\midrule
Conclusion wrong & $n_{\mathrm{TP}}$ (true positive) & $n_{\mathrm{FN}}$ (false negative) & \multirow{2}{*}{$n_{\mathrm{U}}$} \\
Conclusion right & $n_{\mathrm{FP}}$ (false positive) & $n_{\mathrm{TN}}$ (true negative) & \\
\bottomrule
\end{tabular}}
\label{tab:confusion}
\end{table}

\begin{table}
\tbl{Metrics defined on Table~\ref{tab:confusion}. Arrows: larger ($\uparrow$) or smaller ($\downarrow$) is better.}
{\begin{tabularx}{\textwidth}{@{}>{\raggedright\arraybackslash}p{0.17\textwidth} l >{\raggedright\arraybackslash}X >{\raggedright\arraybackslash}p{0.25\textwidth}@{}}
\toprule
Metric & Symbol & Definition & Parameters \\
\midrule
Coverage $\uparrow$ & $\gamma$ & $1-n_{\mathrm{U}}/N$, the share of reasonings a method decides & -- \\
False positives $\downarrow$ & FP & $n_{\mathrm{FP}}/(n_{\mathrm{TP}}+n_{\mathrm{FP}})$, the share of rejections whose conclusion is right (false discovery rate) & -- \\
False negatives $\downarrow$ & FN & $n_{\mathrm{FN}}/(n_{\mathrm{TN}}+n_{\mathrm{FN}})$, the share of acceptances whose conclusion is wrong (false omission rate) & -- \\
F1 on covered cases $\uparrow$ & F1 & $2n_{\mathrm{TP}}/(2n_{\mathrm{TP}}+n_{\mathrm{FP}}+n_{\mathrm{FN}})$, the harmonic mean of precision $n_{\mathrm{TP}}/(n_{\mathrm{TP}}+n_{\mathrm{FP}})$ and recall $n_{\mathrm{TP}}/(n_{\mathrm{TP}}+n_{\mathrm{FN}})$ & -- \\
Effective reliability $\uparrow$ & $\Phi_c$ & $\bigl(n_{\mathrm{TP}}+n_{\mathrm{TN}}-c\,(n_{\mathrm{FP}}+n_{\mathrm{FN}})\bigr)/N$ & $c$: penalty for a wrong decision (a right decision scores 1) \\
Expected cost $\downarrow$ & $C_{R,L}$ & $p + R\,(n_{\mathrm{TP}}+n_{\mathrm{FP}}+n_{\mathrm{U}})/N + L\,n_{\mathrm{FN}}/N$, in US dollars per reasoning & $p$: LLM verification cost; $R$: human review cost; $L$: false negative penalty \\
\bottomrule
\end{tabularx}}
\label{tab:metrics}
\end{table}

\paragraph{Coverage and errors} A verifier is useful only if it decides, and correct only if its decisions are right. Coverage $\gamma$ is the share of reasonings a method decides, measuring usefulness. FP, FN and F1 are quantities that measure correctness, computed on the decided cases only. FN is the error that matters most: a false negative is a wrong reasoning that is mistakenly trusted.

\paragraph{Effective reliability} Coverage and correctness usually trade off: a method that decides more cases risks more errors; one that never decides makes zero errors, while being useless. Therefore, neither ranks verification reliability alone. To weigh usefulness (coverage) against correctness (errors), effective reliability $\Phi_c$ \citep{whitehead2022reliable} scores all $N$ reasonings in one number: a right decision scores 1, a wrong decision $-c$, and a reasoning the method cannot decide 0. A method that does not decide at all scores 0, being a baseline for all verifiers. A verifier is only worthy when its $\Phi_c \gg 0$, i.e. it decides enough cases with an adequate correct rate.

\paragraph{Expected cost}\label{sec:metric-dollar} $\Phi_c$ only counts the weighted importance of right and wrong decisions. However, making decisions itself costs time an money. If finding a wrong case costs way more than the problem the error may cause, it is economically not worth verification at all. Besides that, refusing to make decisions also costs money -- a case that a verifier cannot decide eventually goes to human review, which is often much more expensive than automated verification. Therefore, $C_{R,L}$ is defined as the weighted sum of non-decision cost (i.e., how much getting a manual review costs), verification cost $R$ (i.e., how much getting an LLM verification costs), and error cost $L$ (i.e., how much a wrongly accepted error costs) \citep{chow1970optimum,elkan2001foundations,drummond2006cost}.


In our evaluation, $p$ is the actual API expense in dollars for each LLM model we use. $R$ and $L$ reflects how ``serious'' the applications are. For daily, personal conversational use, a mistake does not matter that much and getting a manual review is cheap. We tentatively set $R=4$ based on the estimate that a person checks a conclusion on a map in about ten minutes at the US median wage of \$23.80 an hour \citep{bls2024oews}, and $L=40$, i.e. a wrong reasoning costs 10 times as a manual review. For general business use, we set $R=20$ based on the estimate that a GIS analyst reviews a reasoning in about twenty minutes at the median pay of cartographers, \$37.68 an hour \citep{bls2024carto}, and $L=1000$. Finally, for high stake cases, such as disaster relief or emergency response, we set $R=100$ and $L=10000$, punishing mistakes heavily. In Figure \ref{fig:cost-sensitivity}, we show that our evaluation is not sensitive to what specific $R$ and $L$ we choose, because \gd{} outperforms the baselines in most cases.

\subsection{Main results}\label{sec:res-detection}
This subsection answers Q1 and Q2. We first compare how reliably each method tells wrong conclusions from right ones on all 585 reasonings (Section~\ref{sec:res-main}). We then test whether a method can locate where an error lies in a reasoning, and whether it can distinguish wrong reasonings from wrong conclusions (Section~\ref{sec:res-injection}). Finally, we use the verification results to assist LLMs fixing wrong reasonings, which is critical for a self-improving GIS pipeline (Section~\ref{sec:res-repair}).

\subsubsection{Verification reliability and cost}\label{sec:res-main}
Tables~\ref{tab:detection} and~\ref{tab:detection-small} compare the verification reliability between \gd{} and existing baselines comprehensively. Except for coverage, \gd{} outperforms all baselines by large margins, regardless of which LLM model is used.

We can see that the baselines trade coverage for F1. On the large tier, re-solve or judge baselines cover 95--100\% of the cases, compared with DATUM's 65.3--86.7\%, yet their F1 range only from 0.56--0.76, compared with DATUM's 0.87--0.95. Combining re-solve and judge baselines improves F1 to 0.69--0.81, still significantly lower than DATUM's, but its coverage even drops below DATUM's. The observation is that the LLM baselines are too eager to give a definite answer even if it is not sure about the correctness (sometimes it hallucinates place names and coordinates, just to reach a conclusion), which results in useless high coverage. This is often referred to as LLM sycophancy \citep{fanous2025syceval,laban2023flipflop,perez2023discovering,sharma2024sycophancy,turpin2023unfaithful,wei2023sycophancy}.

\begin{table}
\tbl{Main results for large-tier models. Jury baselines are put at the bottom because they combine judges from three LLM providers. Coverage (\%), FP, FN, F1 on covered cases, $\Phi_c$ and $C_{R,L}$ are defined in Section~\ref{sec:metrics}; $\Phi_c$ is given at $c=3$ and 10, and $C_{R,L}$ at $(R,L)=(4,40)$, $(20,1000)$ and $(100,10000)$ (Section~\ref{sec:metric-dollar}). Expense: the API expense, dollars actually spent on model calls. A dash marks a value that does not apply. Arrows: larger ($\uparrow$) or smaller ($\downarrow$) is better. Bold: best per model and column.}
{\begin{tabular}{l rr rrr rr rrr}
\toprule
 & & & & & & & & \multicolumn{3}{c}{Expected cost $C_{R,L}$ (\$)} \\
\cmidrule(lr){9-11}
Method & Coverage & Expense (\$) & FP $\downarrow$ & FN $\downarrow$ & F1 $\uparrow$ & $\Phi_3$ $\uparrow$ & $\Phi_{10}$ $\uparrow$ & $C_{4,40}$ $\downarrow$ & $C_{20,1000}$ $\downarrow$ & $C_{100,10000}$ $\downarrow$ \\
\midrule
\multicolumn{11}{@{}l}{\emph{No verifier}}\\
Review everything & 0.0 & -- & -- & -- & -- & 0.00 & 0.00 & 4.00 & 20.0 & 100 \\
Accept everything & 100.0 & -- & -- & -- & -- & $-$0.08 & $-$1.97 & 10.8 & 270 & 2701 \\
\midrule
\multicolumn{11}{@{}l}{\emph{GPT-6 Sol}}\\
Re-solve & \textbf{99.7} & 3.24 & 0.337 & 0.081 & 0.72 & 0.34 & $-$0.81 & 3.51 & 61.3 & 580 \\
\addlinespace
Generic judge & 99.5 & \textbf{2.41} & 0.420 & 0.135 & 0.61 & 0.12 & $-$1.41 & 4.98 & 100 & 970 \\
Informed judge & 97.8 & 3.12 & 0.408 & 0.087 & 0.67 & 0.22 & $-$1.11 & 3.66 & 63.4 & 599 \\
Step-level judge & 95.4 & 5.01 & 0.421 & 0.075 & 0.67 & 0.20 & $-$1.11 & 3.39 & 53.8 & 500 \\
\addlinespace
Re-solve + generic judge & 81.7 & 5.65 & 0.271 & 0.072 & 0.76 & 0.41 & $-$0.31 & 3.33 & 50.8 & 468 \\
Re-solve + informed judge & 82.1 & 6.36 & 0.264 & 0.056 & 0.79 & 0.44 & $-$0.23 & 2.99 & 40.9 & 367 \\
Re-solve + step-level judge & 78.8 & 8.25 & 0.245 & 0.053 & 0.80 & 0.44 & $-$0.17 & 2.97 & 38.1 & 336 \\
\addlinespace
\gd{} (ours) & 86.7 & 11.24 & \textbf{0.068} & \textbf{0.011} & \textbf{0.95} & \textbf{0.78} & \textbf{0.62} & \textbf{1.74} & \textbf{14.1} & \textbf{104} \\
\midrule
\multicolumn{11}{@{}l}{\emph{Claude Sonnet 5.5}}\\
Re-solve & \textbf{99.7} & 5.15 & 0.500 & 0.150 & 0.56 & $-$0.08 & $-$1.97 & 5.30 & 104 & 1009 \\
\addlinespace
Generic judge & 98.8 & \textbf{4.30} & 0.477 & 0.103 & 0.62 & 0.02 & $-$1.68 & 4.09 & 71.0 & 671 \\
Informed judge & 99.1 & 5.75 & 0.547 & 0.077 & 0.59 & $-$0.27 & $-$2.47 & 3.57 & 47.9 & 428 \\
Step-level judge & 97.4 & 8.21 & 0.414 & 0.098 & 0.66 & 0.18 & $-$1.21 & 3.95 & 70.3 & 668 \\
\addlinespace
Re-solve + generic judge & 74.0 & 9.46 & 0.387 & 0.078 & 0.69 & 0.22 & $-$0.69 & 3.56 & 49.2 & 443 \\
Re-solve + informed judge & 69.9 & 10.91 & 0.426 & 0.065 & 0.69 & 0.12 & $-$0.90 & 3.42 & 38.9 & 331 \\
Re-solve + step-level judge & 73.3 & 13.36 & 0.328 & 0.078 & 0.72 & 0.30 & $-$0.47 & 3.56 & 50.6 & 458 \\
\addlinespace
\gd{} (ours) & 65.3 & 23.75 & \textbf{0.195} & \textbf{0.019} & \textbf{0.87} & \textbf{0.45} & \textbf{0.11} & \textbf{2.61} & \textbf{19.7} & \textbf{141} \\
\midrule
\multicolumn{11}{@{}l}{\emph{Gemini 4 Argon}}\\
Re-solve & \textbf{99.8} & 5.06 & 0.291 & 0.072 & 0.76 & 0.44 & $-$0.54 & 3.24 & 55.8 & 527 \\
\addlinespace
Generic judge & 99.7 & \textbf{3.49} & 0.245 & 0.117 & 0.71 & 0.41 & $-$0.62 & 4.53 & 93.7 & 913 \\
Informed judge & 98.8 & 4.40 & 0.354 & 0.090 & 0.70 & 0.31 & $-$0.87 & 3.73 & 67.9 & 647 \\
Step-level judge & 97.8 & 7.01 & 0.250 & 0.093 & 0.74 & 0.46 & $-$0.45 & 3.82 & 73.8 & 711 \\
\addlinespace
Re-solve + generic judge & 87.5 & 8.55 & 0.194 & 0.062 & 0.81 & 0.55 & $-$0.03 & 3.00 & 47.8 & 444 \\
Re-solve + informed judge & 85.5 & 9.46 & 0.232 & 0.052 & 0.81 & 0.51 & $-$0.10 & 2.84 & 40.1 & 363 \\
Re-solve + step-level judge & 86.8 & 12.07 & 0.206 & 0.052 & 0.81 & 0.55 & 0.00 & 2.78 & 41.1 & 377 \\
\addlinespace
\gd{} (ours) & 85.0 & 23.98 & \textbf{0.054} & \textbf{0.016} & \textbf{0.95} & \textbf{0.76} & \textbf{0.61} & \textbf{1.94} & \textbf{17.8} & \textbf{140} \\
\midrule
\multicolumn{11}{@{}l}{\emph{Juries: majority of the same judge on the three models above}}\\
Generic judges & 99.1 & 10.20 & 0.371 & 0.116 & 0.66 & 0.24 & $-$1.08 & 4.48 & 87.9 & 850 \\
Informed judges & 98.3 & 13.27 & 0.422 & 0.081 & 0.67 & 0.18 & $-$1.22 & 3.55 & 58.7 & 550 \\
Step-level judges & 97.4 & 20.23 & 0.373 & 0.092 & 0.69 & 0.28 & $-$0.94 & 3.81 & 68.1 & 648 \\
\bottomrule
\end{tabular}}
\label{tab:detection}
\end{table}

\begin{table}
\tbl{Main results for small-tier models. Setup as in Table~\ref{tab:detection}.}
{\begin{tabular}{l rr rrr rr rrr}
\toprule
 & & & & & & & & \multicolumn{3}{c}{Expected cost $C_{R,L}$ (\$)} \\
\cmidrule(lr){9-11}
Method & Coverage & Expense (\$) & FP $\downarrow$ & FN $\downarrow$ & F1 $\uparrow$ & $\Phi_3$ $\uparrow$ & $\Phi_{10}$ $\uparrow$ & $C_{4,40}$ $\downarrow$ & $C_{20,1000}$ $\downarrow$ & $C_{100,10000}$ $\downarrow$ \\
\midrule
\multicolumn{11}{@{}l}{\emph{No verifier}}\\
Review everything & 0.0 & -- & -- & -- & -- & 0.00 & 0.00 & 4.00 & 20.0 & 100 \\
Accept everything & 100.0 & -- & -- & -- & -- & $-$0.08 & $-$1.97 & 10.8 & 270 & 2701 \\
\midrule
\multicolumn{11}{@{}l}{\emph{GPT-6 Luna}}\\
Re-solve & 99.3 & 0.16 & 0.556 & 0.162 & 0.52 & $-$0.23 & $-$2.37 & 5.54 & 108 & 1046 \\
\addlinespace
Generic judge & \textbf{99.8} & \textbf{0.13} & 0.498 & 0.137 & 0.58 & $-$0.07 & $-$1.94 & 4.94 & 94.4 & 908 \\
Informed judge & 97.6 & 0.16 & 0.504 & 0.094 & 0.61 & $-$0.07 & $-$1.90 & 3.87 & 61.7 & 574 \\
Step-level judge & 91.3 & 0.25 & 0.521 & 0.113 & 0.57 & $-$0.03 & $-$1.68 & 4.31 & 74.9 & 708 \\
\addlinespace
Re-solve + generic judge & 69.4 & 0.29 & 0.411 & 0.103 & 0.65 & 0.15 & $-$0.81 & 4.06 & 59.9 & 548 \\
Re-solve + informed judge & 66.8 & 0.32 & 0.400 & 0.086 & 0.68 & 0.15 & $-$0.76 & 3.75 & 48.9 & 432 \\
Re-solve + step-level judge & 60.0 & 0.41 & 0.384 & 0.091 & 0.67 & 0.18 & $-$0.55 & 3.85 & 50.7 & 450 \\
\addlinespace
\gd{} (ours) & 73.0 & 0.55 & \textbf{0.149} & \textbf{0.019} & \textbf{0.89} & \textbf{0.57} & \textbf{0.30} & \textbf{2.27} & \textbf{19.6} & \textbf{149} \\
\midrule
\multicolumn{11}{@{}l}{\emph{Claude Haiku 5.5}}\\
Re-solve & \textbf{99.7} & 0.26 & 0.643 & 0.190 & 0.46 & $-$0.62 & $-$3.44 & 5.87 & 109 & 1039 \\
\addlinespace
Generic judge & 93.2 & \textbf{0.22} & 0.541 & 0.115 & 0.56 & $-$0.09 & $-$1.89 & 4.34 & 75.0 & 709 \\
Informed judge & 95.6 & 0.30 & 0.625 & 0.064 & 0.53 & $-$0.65 & $-$3.46 & 3.49 & 35.2 & 287 \\
Step-level judge & 87.0 & 0.48 & 0.424 & 0.076 & 0.66 & 0.21 & $-$0.94 & 3.43 & 52.7 & 486 \\
\addlinespace
Re-solve + generic judge & 62.4 & 0.47 & 0.518 & 0.096 & 0.59 & $-$0.01 & $-$1.12 & 3.95 & 49.8 & 437 \\
Re-solve + informed judge & 61.7 & 0.56 & 0.587 & 0.052 & 0.57 & $-$0.27 & $-$1.83 & 3.50 & 28.4 & 211 \\
Re-solve + step-level judge & 57.8 & 0.74 & 0.414 & 0.059 & 0.69 & 0.16 & $-$0.57 & 3.37 & 34.6 & 284 \\
\addlinespace
\gd{} (ours) & 71.6 & 1.50 & \textbf{0.185} & \textbf{0.014} & \textbf{0.88} & \textbf{0.52} & \textbf{0.19} & \textbf{2.30} & \textbf{17.0} & \textbf{119} \\
\midrule
\multicolumn{11}{@{}l}{\emph{Gemini 3.1 Flash-Lite}}\\
Re-solve & 99.8 & 0.45 & 0.510 & 0.151 & 0.55 & $-$0.11 & $-$2.05 & 5.31 & 105 & 1010 \\
\addlinespace
Generic judge & \textbf{100.0} & \textbf{0.34} & 0.301 & 0.209 & 0.44 & 0.12 & $-$1.43 & 7.82 & 185 & 1842 \\
Informed judge & 98.8 & 0.44 & 0.434 & 0.163 & 0.55 & 0.08 & $-$1.51 & 5.88 & 126 & 1239 \\
Step-level judge & 99.0 & 0.88 & 0.397 & 0.168 & 0.55 & 0.13 & $-$1.38 & 6.10 & 134 & 1322 \\
\addlinespace
Re-solve + generic judge & 69.9 & 0.79 & 0.250 & 0.126 & 0.57 & 0.30 & $-$0.39 & 4.64 & 84.7 & 808 \\
Re-solve + informed judge & 71.3 & 0.88 & 0.359 & 0.102 & 0.64 & 0.26 & $-$0.53 & 4.04 & 65.3 & 609 \\
Re-solve + step-level judge & 69.4 & 1.33 & 0.282 & 0.107 & 0.66 & 0.30 & $-$0.38 & 4.15 & 68.6 & 642 \\
\addlinespace
\gd{} (ours) & 61.4 & 1.87 & \textbf{0.176} & \textbf{0.044} & \textbf{0.86} & \textbf{0.41} & \textbf{0.05} & \textbf{3.04} & \textbf{30.2} & \textbf{245} \\
\midrule
\multicolumn{11}{@{}l}{\emph{Juries: majority of the same judge on the three models above}}\\
Generic judges & 96.6 & 0.68 & 0.433 & 0.140 & 0.59 & 0.11 & $-$1.38 & 5.11 & 103 & 1005 \\
Informed judges & 96.9 & 0.90 & 0.510 & 0.098 & 0.60 & $-$0.08 & $-$1.93 & 3.95 & 63.5 & 591 \\
Step-level judges & 91.1 & 1.61 & 0.392 & 0.106 & 0.65 & 0.23 & $-$0.95 & 4.17 & 76.9 & 735 \\
\bottomrule
\end{tabular}}
\label{tab:detection-small}
\end{table}

Coverage and F1 alone do not fully rank the methods: re-solve or a single judge decides nearly everything at a lower F1, while a re-solve + judge baseline or \gd{} decides fewer cases at a higher one. Effective reliability gives them a weighted ranking. At $\Phi_3$, i.e. a wrong verification matters three times as a correct one, \gd{} is highest on every model: 0.78, 0.45 and 0.76 on the large tier and 0.57, 0.52 and 0.41 on the small tier, against at most 0.44, 0.22, 0.55, 0.15, 0.21 and 0.30 for the baselines. At $\Phi_{10}$, all baselines score below zero, meaning that using a baseline verifier is even worse than not reviewing at all. In the contrast, \gd{} remains positive on all models, showing its consistent usefulness and reliability.

In dollars, \gd{}'s API expense reaches \$0.55--24 (\$11.24 on GPT-6 Sol), more than any single baseline's, but still extremely cheap compared to the cost of human reviews and error risks. In daily use, \gd{} is the cheapest policy on every model: $C_{4,40}$ is \$1.74--3.04 per reasoning, against \$2.84--4.04 for the cheapest baseline, \$4 for reviewing everything and \$10.80 for accepting everything. In business use, every baseline, the juries included, costs more than reviewing everything (at least \$28.40 against \$20), while \gd{} costs \$14.10--19.70 on five models and \$30.20 on Flash-Lite. In serious cases, reviewing everything is cheapest (\$100); \gd{} comes closest (\$104 on GPT-6 Sol, \$119--245 on the other models), and the baselines cost at least \$211. By the break-even ratios (Section~\ref{sec:res-cost-sensitivity}), \gd{} is cheaper than full review while a false negative costs less than about 23--93 reviews, the baselines only below 10--19: \gd{} widens, by two to five times, the range of stakes at which an LLM's spatial conclusion can be trusted without a person.

\begin{figure}
\centering
\includegraphics[width=\textwidth]{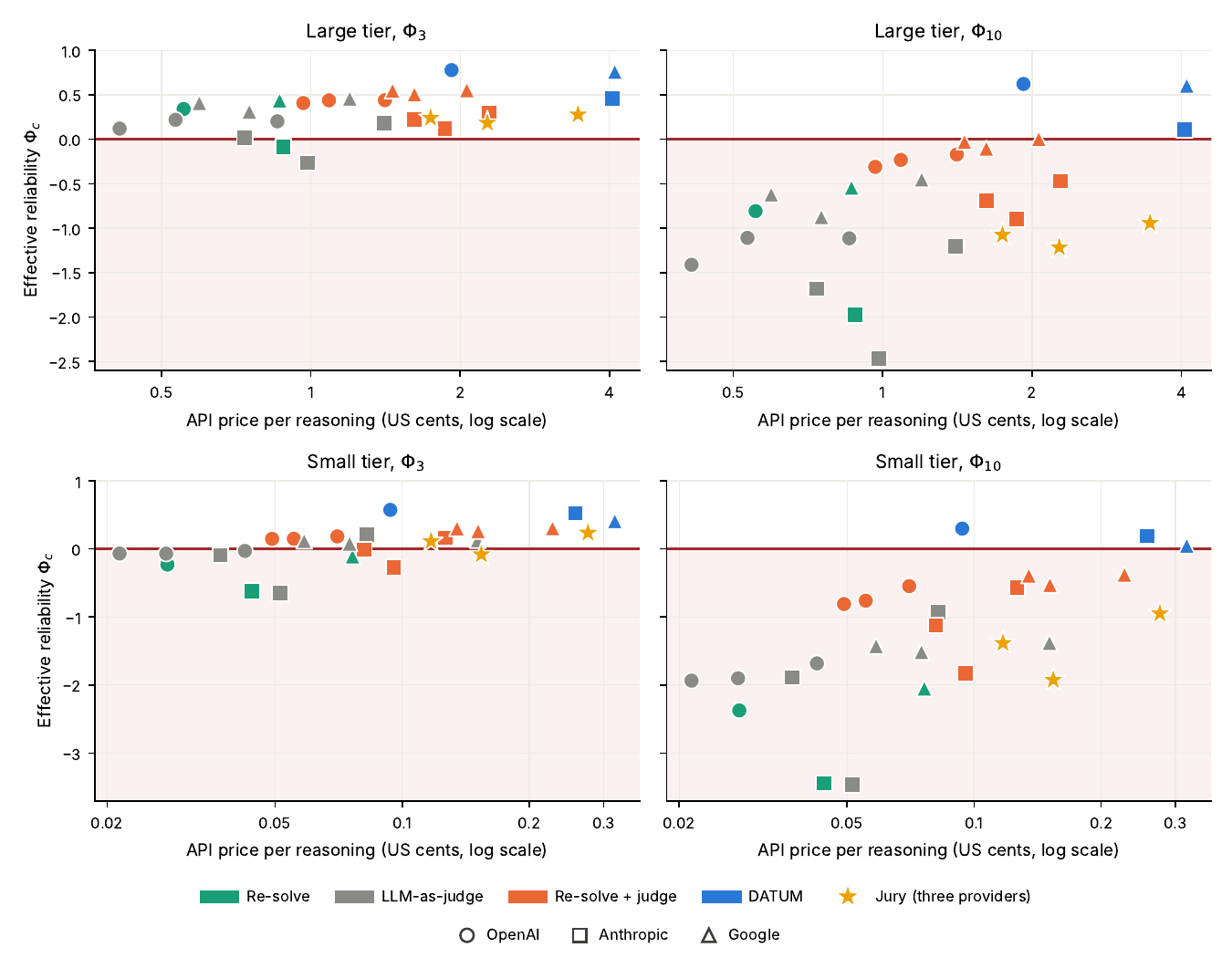}
\caption{Effective reliability against API price per reasoning (log scale). One point per method and model. Color: baseline group (re-solve; LLM-as-judge; re-solve + judge) or \gd{}; shape: model provider. Stars: juries, each one judge over the tier's three providers. The shaded band ($\Phi_c<0$) is worse than deciding no case.}
\label{fig:price-phi}
\end{figure}

Figure~\ref{fig:price-phi} sets each method's effective reliability against its API price per reasoning. On every model \gd{} is the highest point at both $\Phi_3$ and $\Phi_{10}$, and the only one above zero at $\Phi_{10}$.

\subsubsection{Error localization}\label{sec:res-injection}\label{sec:res-localization}
Conclusion correctness cannot tell whether a report names the right cause, nor whether a method can distinguish a wrong reasoning from a wrong conclusion. We construct a control dataset by planting an error in a statement in each of the 283 correct reasonings and evaluate the verifiers on it. We first compare whether each verifier locates the actual wrong statement accurately, and then compares if the verifiers change their mind on whether the conclusion is correct (notice: we only changed a statement; the answers of these reasonings are still correct). Specifically, we use a fixed script to edit one value at a statement in six different ways (Table~\ref{tab:injection}).

\paragraph{Locating planted errors} \gd{} marks 254 planted errors out of 283 (90\%). The step-level judge, which labels every statements separately, reaches 229 (81\%); the informed judge, which verifies the reasoning as a whole, only reaches 50 (18\%). A generic judge only decides whether the conclusions are correct or wrong, so it is left out.

Of the 29 (10\%) planted errors \gd{} misses, 11 come from that the edited values no longer match the requirements of the typed contracts, so they are immediately rejected without error localization. 12 are distance edits whose effect on the conclusion is smaller than the 0.5~km tolerance, which technically are not errors. Only 6 cases are truly missed errors.

\paragraph{Telling a flawed step from a wrong conclusion} We use this experiment to test whether a verifier distinguishes a flawed reasoning step from a wrong conclusion. This is important because LLMs hallucinate -- they sometimes give a correct conclusion with wrong reasoning, and sometime the other way around. Since the controlled dataset we construct consists of wrong reasonings and correct conclusion, a natural metric is how many conclusions are labeled ``wrong'' -- called \emph{false rejection} -- when the verifier is presented a wrong reasoning. 

\gd{} has zero false rejections: it reports each planted error yet accepts the conclusions, because \gd{} has a re-computation module in its kernel. The informed judge has 72 false rejections (25\%) and the step-level judge 68 (24\%). This shows that locating an error and judging its consequence are separate skills: the step-level judge finds the planted errors nearly as often as \gd{} (90\% v.s. 81\%), but works way worse in verifying conclusions independently.

\begin{table}
\tbl{Error localization on planted errors in 283 reasonings with right conclusions. \emph{Error located}: \gd{} marks the planted statement inconsistent; a judge names the planted sentence as the first wrong one. \emph{False rejection}: the method rejects the conclusion, which is right in every case. Arrows: whether a larger or smaller number is better.}
{\begin{tabular}{lrrrrrrr}
\toprule
 & & \multicolumn{3}{c}{Error located $\uparrow$} & \multicolumn{3}{c}{False rejection $\downarrow$} \\
\cmidrule(lr){3-5}\cmidrule(lr){6-8}
Planted error & Cases & \gd{} & Informed & Step-level & \gd{} & Informed & Step-level \\
\midrule
Distance $\times$1.5 & 63 & 47 & 13 & 50 & 0 & 19 & 21 \\
Travel time $\times$1.5 & 42 & 41 & 1 & 41 & 0 & 2 & 1 \\
Arithmetic result & 26 & 25 & 4 & 22 & 0 & 6 & 2 \\
Clock time +1 h & 14 & 11 & 2 & 10 & 0 & 3 & 0 \\
Latitude +0.3$^\circ$ & 75 & 71 & 21 & 68 & 0 & 26 & 25 \\
Compass direction flipped & 63 & 59 & 9 & 38 & 0 & 16 & 19 \\
\midrule
All & 283 & 254 & 50 & 229 & 0 & 72 & 68 \\
Of 283 cases & & 90\% & 18\% & 81\% & 0\% & 25\% & 24\% \\
\bottomrule
\end{tabular}}
\label{tab:injection}
\end{table}

\subsubsection{Repair and re-verification}\label{sec:res-repair}
An LLM rarely corrects its own reasoning without outside feedback \citep{huang2024selfcorrect}. Self-correction works when reliable external feedback is available \citep{kamoi2024selfcorrection}, such as the output of a tool \citep{gou2024critic} or a sound verifier \citep{kambhampati2024llmmodulo}. For an autonomous GIS pipeline where the in-between steps are less often manually checked \citep{li2023autonomous}, a self-correction is critical to make the system robust. We test whether a verifier's judgment can be used to repair a wrong reasoning and whether the repaired reasoning can be trusted.

For each method, we feed the original task, context and reasoning together with the verification output (e.g. \gd{} outputs a report) back to GPT-4o, the model that generates the Spatial CoT reasonings. It is required to take the critiques into consideration and re-solve the problem accordingly. \gd{}'s report fixes 103 of the 124 wrong cases, against 84 (informed judge), 83 (step-level judge) and 72 (generic judge) (Table~\ref{tab:repair}). It shows that locating an error does not mean fixing it: the step-level judge, which finds planted errors only 9 fewer than \gd{}, fixes 20 fewer. \gd{}'s lead is largest on GS-QA, whose conclusions are open-ended names, distances and counts (78 of 92, against 61 on informed judge); on MapEval, where the author chooses among four options, the step-level judge's critiques fix as many (25 of 32).

\begin{figure}
\centering
\includegraphics[width=\textwidth]{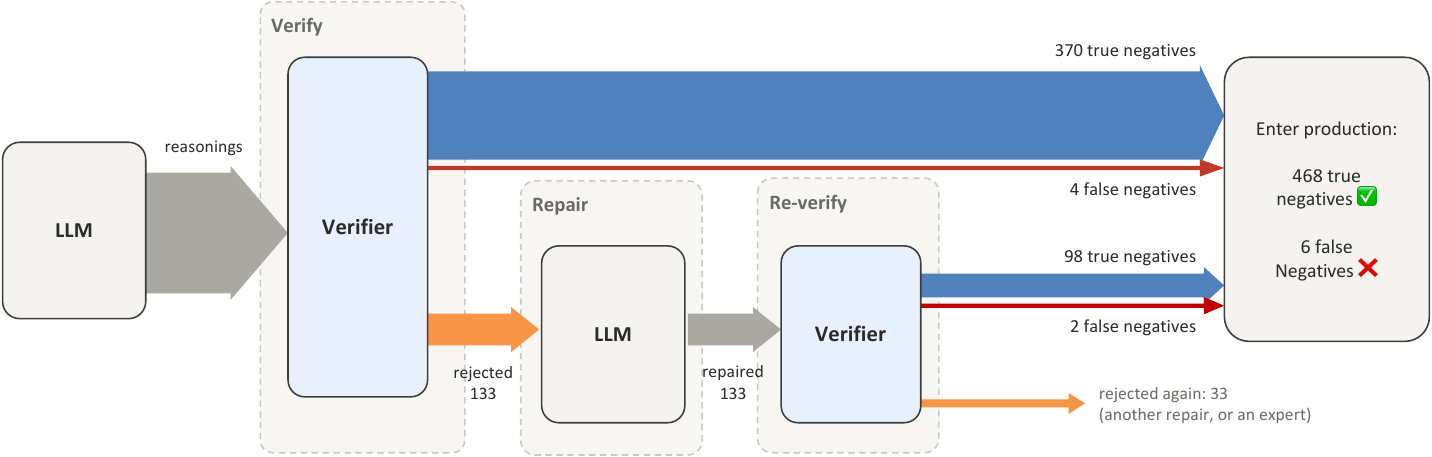}
\caption{The verify--repair--re-verify loop on DATUM with GPT-6 Sol. \emph{Verify} (the first check) passes what it accepts into production and sends what it rejects back to the LLM for \emph{repair} with the verifier's feedback. \emph{Re-verify} (the re-check, by the same verifier) adds 98 true negatives and 2 false negatives. The 33 it rejects again can be repaired once more or passed to an expert.}
\label{fig:repair-loop}
\end{figure}

Repairing is not risk-free. As Figure~\ref{fig:repair-loop} shows: In the first verification round of DATUM on GPT-6 Sol, 4 false negatives enter production, 133 cases are rejected. Running a repair round reduces rejected cases to 33. However, 2 more false negatives pass the re-verification, i.e. DATUM falsely accepts 2 repaired cases that remain wrong. This re-verification false negative rate is critical for a self-correction mechanism to work in an agentic pipeline -- in Table~\ref{tab:repair}, the baseline LLM-as-judge methods introduce dozens more false negatives into production after one round of repair, which almost offsets the gain of repairing. DATUM, which only introduces two more false negatives, makes self-correction practical.

\begin{table}
\tbl{Repair and re-verification false negative rates.}
{\begin{tabular}{lrrrrrr}
\toprule
 & \multicolumn{3}{c}{Wrong conclusions fixed $\uparrow$} & \multicolumn{3}{c}{False negatives $\downarrow$} \\ \cmidrule(lr){2-4} \cmidrule(lr){5-7}
Feedback & MapEval & GS-QA & All & First check & Re-check & All \\
 & (of 32) & (of 92) & (of 124) & & & (of 585) \\
\midrule
\gd{} & 25 & 78 & 103 & 4 & 2 & 6 \\
Informed judge & 23 & 61 & 84 & 33 & 24 & 57 \\
Step-level judge & 25 & 58 & 83 & 27 & 19 & 46 \\
Generic judge & 23 & 49 & 72 & 55 & 22 & 77 \\
\bottomrule
\end{tabular}}
\label{tab:repair}
\end{table}

\subsection{Ablations}\label{sec:res-ablation}
The ablations answer Q3 and test the choices behind the main results: whether the results change with LLM hidden reasoning toggle on (Section~\ref{sec:abl-reasoning}), with self-consistency re-solve (Section~\ref{sec:abl-repeated}), with different LLMs that bind or transcribe (Section~\ref{sec:abl-binder}), without one of the kernel's module (Section~\ref{sec:abl-kernel}), or with different $R$ and $L$ for computing expected cost (Section~\ref{sec:res-cost-sensitivity}).

\subsubsection{Hidden reasoning}\label{sec:abl-reasoning}
Many current LLMs can utilize a hidden chain of thought. The user does not see this \emph{hidden reasoning}, but pays for its tokens as output, and a setting called the reasoning effort (here \emph{off} or \emph{on}) controls how much of it the model writes. All results so far use reasoning off. Table~\ref{tab:onoff} turns it on, at high effort, for every method. It raises F1 only by 0.01--0.08, but the API expense rises by 23\% to 101\%.

\begin{table}
\tbl{Reasoning off against on (high effort) on GPT-6 Sol. Methods and columns as in Table~\ref{tab:detection}. Expense: API expense as in Table~\ref{tab:detection}, with hidden reasoning tokens billed as output. $\Delta$Expense: change from reasoning off.}
{\begin{tabular}{l rrrrr rrrrr rr}
\toprule
 & \multicolumn{5}{c}{Reasoning off} & \multicolumn{5}{c}{Reasoning on (high)} & & \\
\cmidrule(lr){2-6}\cmidrule(lr){7-11}
Method & Coverage & FP $\downarrow$ & FN $\downarrow$ & F1 $\uparrow$ & Expense (\$) & Coverage & FP $\downarrow$ & FN $\downarrow$ & F1 $\uparrow$ & Expense (\$) & $\Delta$F1 & $\Delta$Expense \\
\midrule
Re-solve & 99.7 & 0.337 & 0.081 & 0.72 & 3.24 & 99.8 & 0.273 & 0.062 & 0.78 & 3.98 & $+$0.06 & $+$23\% \\
Generic judge & 99.5 & 0.420 & 0.135 & 0.61 & 2.41 & 99.7 & 0.354 & 0.101 & 0.69 & 3.30 & $+$0.08 & $+$37\% \\
Informed judge & 97.8 & 0.408 & 0.087 & 0.67 & 3.12 & 97.4 & 0.370 & 0.082 & 0.70 & 5.11 & $+$0.03 & $+$64\% \\
Step-level judge & 95.4 & 0.421 & 0.075 & 0.67 & 5.01 & 96.9 & 0.365 & 0.075 & 0.71 & 10.06 & $+$0.03 & $+$101\% \\
Re-solve + generic judge & 81.7 & 0.271 & 0.072 & 0.76 & 5.65 & 86.8 & 0.225 & 0.055 & 0.81 & 7.28 & $+$0.05 & $+$29\% \\
Re-solve + informed judge & 82.1 & 0.264 & 0.056 & 0.79 & 6.36 & 84.4 & 0.210 & 0.051 & 0.82 & 9.09 & $+$0.04 & $+$43\% \\
Re-solve + step-level judge & 78.8 & 0.245 & 0.053 & 0.80 & 8.25 & 84.1 & 0.203 & 0.048 & 0.83 & 14.05 & $+$0.03 & $+$70\% \\
Self-consistency, majority & 99.8 & 0.339 & 0.079 & 0.73 & 16.21 & 99.8 & 0.279 & 0.065 & 0.77 & 19.92 & $+$0.05 & $+$23\% \\
Self-consistency, unanimous & 91.3 & 0.293 & 0.065 & 0.76 & 16.21 & 95.0 & 0.234 & 0.057 & 0.81 & 19.92 & $+$0.04 & $+$23\% \\
\gd{} (ours) & 86.7 & 0.068 & 0.011 & 0.95 & 11.24 & 83.1 & 0.047 & 0.011 & 0.96 & 20.65 & $+$0.01 & $+$84\% \\
\bottomrule
\end{tabular}}
\label{tab:onoff}
\end{table}

\subsubsection{Self-consistency re-solve}\label{sec:abl-repeated}
Table~\ref{tab:repeated} compares self-consistency re-solve, i.e. five independent re-solves deciding together, with a single re-solve. It multiplies the API expense by five, from \$3.24 to \$16.21 on GPT-6 Sol and from \$0.16 to \$0.81 on GPT-6 Luna, without improving the results much. Self-consistency by majority reaches very similar F1 scores on GPT-6 Sol (0.73 v.s. 0.72) and GPT-6 Luna (0.56 v.s. 0.52). Self-consistency by unanimity reaches slightly better F1 scores on GPT-6 Sol (0.76 v.s. 0.72) and GPT-6 Luna (0.59 v.s. 0.52), at the cost of much worse coverage on GPT-6 Sol (91.3\% v.s. 99.7\%) and GPT-6 Luna (75.9\% v.s. 99.3\%).

\begin{table}
\tbl{Self-consistency re-solve v.s. single re-solve (Tables~\ref{tab:detection} and~\ref{tab:detection-small}). Columns as in Table~\ref{tab:detection}.}
{\begin{tabular}{l l l l l l l l l l}
\toprule
Method & Coverage & FP $\downarrow$ & FN $\downarrow$ & F1 $\uparrow$ & $\Phi_3$ $\uparrow$ & $\Phi_{10}$ $\uparrow$ & $C_{4,40}$ $\downarrow$ & $C_{20,1000}$ $\downarrow$ & $C_{100,10000}$ $\downarrow$ \\
\midrule
\multicolumn{10}{@{}l}{\emph{Large tier: GPT-6 Sol}}\\
Re-solve, once & 99.7 & 0.337 & 0.081 & 0.72 & 0.34 & $-$0.81 & 3.51 & 61.3 & 580 \\
Self-consistency, majority & 99.8 & 0.339 & 0.079 & 0.73 & 0.34 & $-$0.81 & 3.47 & 59.6 & 563 \\
Self-consistency, unanimous & 91.3 & 0.293 & 0.065 & 0.76 & 0.41 & $-$0.46 & 3.16 & 48.5 & 448 \\
\midrule
\multicolumn{10}{@{}l}{\emph{Small tier: GPT-6 Luna}}\\
Re-solve, once & 99.3 & 0.556 & 0.162 & 0.52 & $-$0.23 & $-$2.37 & 5.54 & 108 & 1046 \\
Self-consistency, majority & 99.1 & 0.533 & 0.133 & 0.56 & $-$0.16 & $-$2.17 & 4.80 & 88.3 & 843 \\
Self-consistency, unanimous & 75.9 & 0.494 & 0.107 & 0.59 & 0.03 & $-$1.25 & 4.14 & 63.1 & 580 \\
\bottomrule
\end{tabular}}
\label{tab:repeated}
\end{table}

\subsubsection{LLM use in \bnd{} and \tsb{}}\label{sec:abl-binder}
\gd{} uses an LLM in two places. The \bnd{} is the main semantic interpreter: it reads the question and states it as a typed goal. The \tsb{} is mainly a parser: it rewrites the reasoning into typed statements, only calling an LLM when encountering unseen data formats. In principle, \bnd{} should rely relatively more on the capability of LLMs than \tsb{}. This subsection, we swap the models used for \bnd{} or \tsb{} while fixing the other at GPT-6 Sol to verify this assertion.

Table~\ref{tab:binder} clearly shows that LLMs for the transcriber barely matter. All models score within 1\% difference. It shows that we can safely use a small-tier LLM for transcribing, which saves money (\$0.10 v.s. \$2.00).

LLMs for the binder do matter. GPT-6 Sol and Gemini 4 Argon show similar results, but Claude Sonnet 5.5 drops from F1 0.95 to F1 0.86, mostly because Claude Sonnet 5.5 does not have as comprehensive support for structured input/output, which our DATUM kernel relies heavily on. Large-tier models do not outperform small-tier models on F1; instead, they have much higher coverage. This means that the verification reliability of DATUM does not rely on LLM's capability of reasoning, because both large and small tier models can verify cases that are successfully bound; the difference is the LLM's skills of binding free-text to given templates, because coverage reflects how many cases are successfully transformed into a structured JSON format. That means we can deploy much smaller local LLMs that are trained for binding, which would greatly reduce the cost of DATUM.

\begin{table}
\tbl{Which LLM module matters in \gd{}. Columns as in Table~\ref{tab:detection}.}
{\begin{tabular}{llllllll}
\toprule
Binder & Transcriber & Coverage (\%) & FP $\downarrow$ & FN $\downarrow$ & F1 $\uparrow$ & $\Phi_3$ $\uparrow$ & $\Phi_{10}$ $\uparrow$ \\
\midrule
\multirow{6}{*}[-0.25em]{GPT-6 Sol} & GPT-6 Sol & 86.7 & 0.068 & 0.011 & 0.95 & 0.78 & 0.62 \\
 & Claude Sonnet 5.5 & 85.8 & 0.063 & 0.011 & 0.95 & 0.78 & 0.63 \\
 & Gemini 4 Argon & 85.6 & 0.048 & 0.011 & 0.96 & 0.79 & 0.67 \\
\addlinespace
 & GPT-6 Luna & 85.3 & 0.065 & 0.011 & 0.95 & 0.77 & 0.63 \\
 & Claude Haiku 5.5 & 85.8 & 0.062 & 0.011 & 0.95 & 0.78 & 0.63 \\
 & Gemini 3.1 Flash-Lite & 84.8 & 0.066 & 0.011 & 0.95 & 0.77 & 0.62 \\
\midrule
GPT-6 Sol & \multirow{6}{*}[-0.25em]{GPT-6 Sol} & 86.7 & 0.068 & 0.011 & 0.95 & 0.78 & 0.62 \\
Claude Sonnet 5.5 &  & 66.5 & 0.209 & 0.019 & 0.86 & 0.45 & 0.06 \\
Gemini 4 Argon &  & 85.6 & 0.060 & 0.016 & 0.95 & 0.76 & 0.59 \\
\addlinespace
GPT-6 Luna &  & 75.4 & 0.164 & 0.019 & 0.89 & 0.57 & 0.25 \\
Claude Haiku 5.5 &  & 72.0 & 0.160 & 0.014 & 0.90 & 0.55 & 0.25 \\
Gemini 3.1 Flash-Lite &  & 65.8 & 0.218 & 0.044 & 0.84 & 0.38 & $-$0.09 \\
\bottomrule
\end{tabular}}
\label{tab:binder}
\end{table}

\subsubsection{Kernel module ablation study}\label{sec:abl-kernel}
We ablate the kernel's deterministic modules (Table~\ref{tab:components}). \emph{Conclusion check only} ignores the closure check and decides from the re-computation alone. \emph{Closure check only} drops the conclusion check. \emph{One interpretation} randomly chooses an interpretation branch for ambiguous cases. \emph{No tolerance} requires all values to exactly match.

Conclusion check alone trades F1 (+0.02) with coverage (-5.0\%), because in the full kernel, the closure check helps decide on cases when re-computation is impossible. Closure check alone shows large F1 drop (-0.28) and coverage drop (-36.1\%), because many Spatial CoT reasonings skip critical steps that make closure check impossible. Single interpretation does not matter that much, because under most ambiguous cases either interpretation works. No tolerance also has minimal influence, because in most cases the correct value is largely different from the wrong options.

\begin{table}
\tbl{Ablation of the kernel's modules. Columns as in Table~\ref{tab:detection}; in parentheses, the difference compared with the full kernel results.}
{\begin{tabular}{l r@{\,}l r@{\,}l r@{\,}l r@{\,}l r@{\,}l r@{\,}l}
\toprule
Variant & \multicolumn{2}{c}{Coverage (\%)} & \multicolumn{2}{c}{FP $\downarrow$} & \multicolumn{2}{c}{FN $\downarrow$} & \multicolumn{2}{c}{F1 $\uparrow$} & \multicolumn{2}{c}{$\Phi_3$ $\uparrow$} & \multicolumn{2}{c}{$\Phi_{10}$ $\uparrow$} \\
\midrule
Full kernel & 86.7 &  & 0.068 &  & 0.011 &  & 0.95 &  & 0.78 &  & 0.62 &  \\
Conclusion check only & 81.7 & {\footnotesize ($-$5.0)} & 0.028 & {\footnotesize ($-$0.040)} & 0.011 & {\footnotesize ($\pm$0.000)} & 0.97 & {\footnotesize ($+$0.02)} & 0.77 & {\footnotesize ($-$0.01)} & 0.69 & {\footnotesize ($+$0.07)} \\
Closure check only & 50.6 & {\footnotesize ($-$36.1)} & 0.487 & {\footnotesize ($+$0.419)} & 0.014 & {\footnotesize ($+$0.003)} & 0.67 & {\footnotesize ($-$0.28)} & $-$0.02 & {\footnotesize ($-$0.80)} & $-$0.94 & {\footnotesize ($-$1.56)} \\
One interpretation & 85.3 & {\footnotesize ($-$1.4)} & 0.089 & {\footnotesize ($+$0.021)} & 0.008 & {\footnotesize ($-$0.003)} & 0.94 & {\footnotesize ($-$0.01)} & 0.75 & {\footnotesize ($-$0.03)} & 0.57 & {\footnotesize ($-$0.05)} \\
No tolerance & 85.1 & {\footnotesize ($-$1.6)} & 0.391 & {\footnotesize ($+$0.323)} & 0.014 & {\footnotesize ($+$0.003)} & 0.75 & {\footnotesize ($-$0.20)} & 0.28 & {\footnotesize ($-$0.50)} & $-$0.71 & {\footnotesize ($-$1.33)} \\
\bottomrule
\end{tabular}}
\label{tab:components}
\end{table}

\subsubsection{Sensitivity of the expected cost}\label{sec:res-cost-sensitivity}
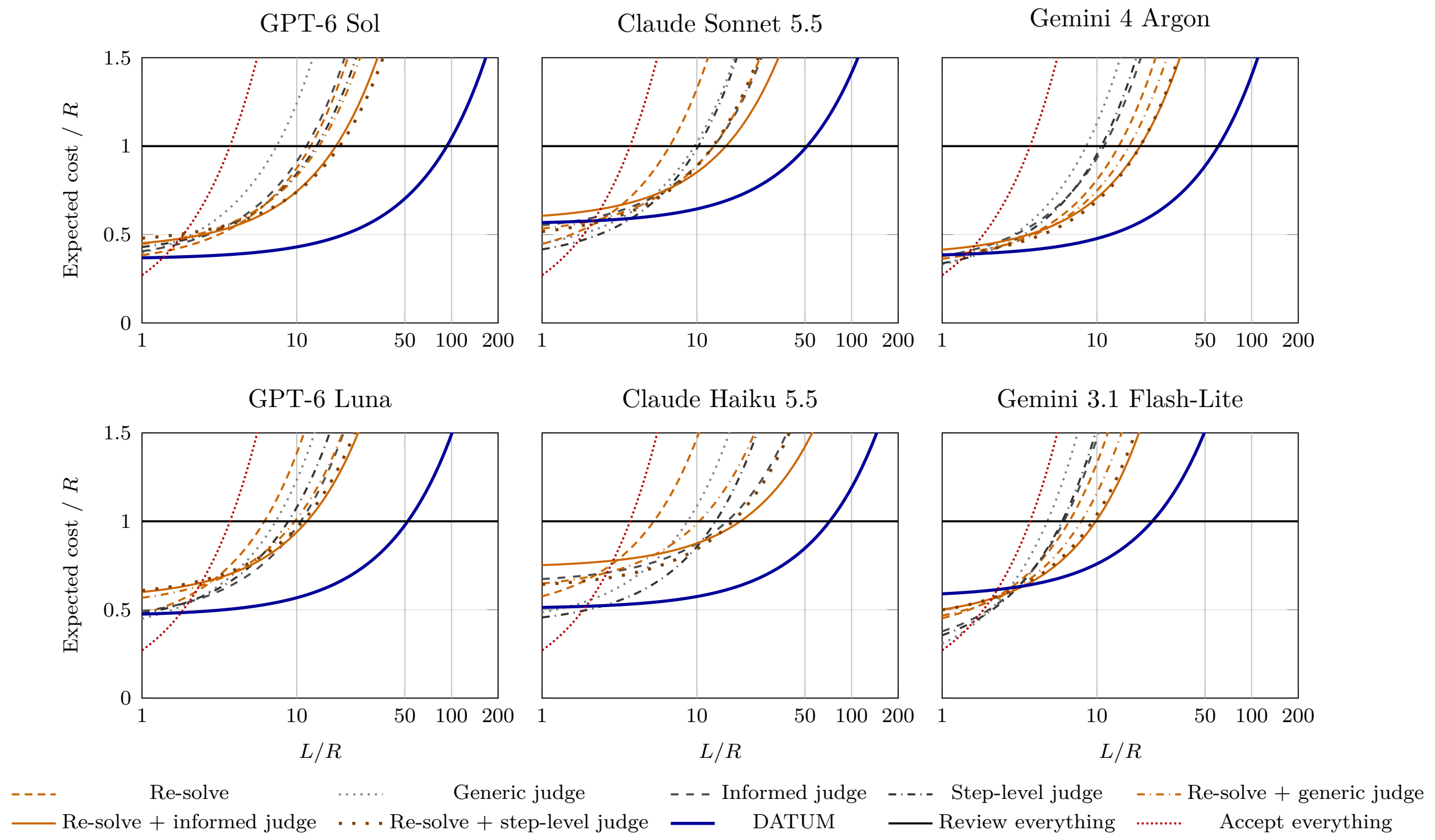
\begin{figure}
\centering
\adjustbox{max width=\textwidth}{
\begin{tikzpicture}
\pgfplotsset{every axis/.style={width=0.39\textwidth, height=5.2cm, xmode=log, xmin=1, xmax=200, ymin=0, ymax=1.5, domain=1:200, samples=80,
  xtick={1,10,50,100,200}, xticklabels={1,10,50,100,200}, ytick={0,0.5,1,1.5}, ymajorgrids, grid style={gray!20}, tick label style={font=\scriptsize},
  label style={font=\scriptsize}, title style={font=\footnotesize}, clip=true}}
\begin{axis}[name=a, title={GPT-6 Sol}, ylabel={Expected cost / $R$}]
\addplot[black!25, thin, forget plot] coordinates {(10,0) (10,1.5)};
\addplot[black!25, thin, forget plot] coordinates {(50,0) (50,1.5)};
\addplot[black!25, thin, forget plot] coordinates {(100,0) (100,1.5)};
\addplot[orange!80!black, densely dashed, thick] {0.3285 + 0.05470*x};
\addplot[gray, dotted, thick] {0.3028 + 0.09402*x};
\addplot[gray!60!black, dashed, thick] {0.3490 + 0.05641*x};
\addplot[gray!40!black, dashdotted, thick] {0.3833 + 0.04615*x};
\addplot[orange!80!black, dashdotted, thick] {0.4039 + 0.04274*x};
\addplot[orange!80!black, thick] {0.4193 + 0.03248*x};
\addplot[orange!50!black, loosely dotted, very thick] {0.4503 + 0.02906*x};
\addplot[blue!60!black, very thick] {0.3616 + 0.00684*x};
\addplot[black, thick] {1};
\addplot[red!70!black, thick, densely dotted] {0.27009*x};
\end{axis}
\begin{axis}[name=b, at={(a.east)}, anchor=west, xshift=0.6cm, title={Claude Sonnet 5.5}, yticklabels={}]
\addplot[black!25, thin, forget plot] coordinates {(10,0) (10,1.5)};
\addplot[black!25, thin, forget plot] coordinates {(50,0) (50,1.5)};
\addplot[black!25, thin, forget plot] coordinates {(100,0) (100,1.5)};
\addplot[orange!80!black, densely dashed, thick] {0.3492 + 0.09744*x};
\addplot[gray, dotted, thick] {0.3884 + 0.06325*x};
\addplot[gray!60!black, dashed, thick] {0.5150 + 0.03761*x};
\addplot[gray!40!black, dashdotted, thick] {0.3528 + 0.06325*x};
\addplot[orange!80!black, dashdotted, thick] {0.4948 + 0.03932*x};
\addplot[orange!80!black, thick] {0.5787 + 0.02735*x};
\addplot[orange!50!black, loosely dotted, very thick] {0.4764 + 0.04103*x};
\addplot[blue!60!black, very thick] {0.5593 + 0.00855*x};
\addplot[black, thick] {1};
\addplot[red!70!black, thick, densely dotted] {0.27009*x};
\end{axis}
\begin{axis}[name=c, at={(b.east)}, anchor=west, xshift=0.6cm, title={Gemini 4 Argon}, yticklabels={}]
\addplot[black!25, thin, forget plot] coordinates {(10,0) (10,1.5)};
\addplot[black!25, thin, forget plot] coordinates {(50,0) (50,1.5)};
\addplot[black!25, thin, forget plot] coordinates {(100,0) (100,1.5)};
\addplot[orange!80!black, densely dashed, thick] {0.3133 + 0.04957*x};
\addplot[gray, dotted, thick] {0.2413 + 0.08889*x};
\addplot[gray!60!black, dashed, thick] {0.3166 + 0.06154*x};
\addplot[gray!40!black, dashdotted, thick] {0.2690 + 0.06838*x};
\addplot[orange!80!black, dashdotted, thick] {0.3375 + 0.04103*x};
\addplot[orange!80!black, thick] {0.3820 + 0.03248*x};
\addplot[orange!50!black, loosely dotted, very thick] {0.3480 + 0.03419*x};
\addplot[blue!60!black, very thick] {0.3747 + 0.01026*x};
\addplot[black, thick] {1};
\addplot[red!70!black, thick, densely dotted] {0.27009*x};
\end{axis}
\begin{axis}[name=d, at={(a.south)}, anchor=north, yshift=-1.5cm, title={GPT-6 Luna}, ylabel={Expected cost / $R$}, xlabel={$L/R$}]
\addplot[black!25, thin, forget plot] coordinates {(10,0) (10,1.5)};
\addplot[black!25, thin, forget plot] coordinates {(50,0) (50,1.5)};
\addplot[black!25, thin, forget plot] coordinates {(100,0) (100,1.5)};
\addplot[orange!80!black, densely dashed, thick] {0.3761 + 0.10085*x};
\addplot[gray, dotted, thick] {0.3624 + 0.08718*x};
\addplot[gray!60!black, dashed, thick] {0.4376 + 0.05299*x};
\addplot[gray!40!black, dashdotted, thick] {0.4120 + 0.06667*x};
\addplot[orange!80!black, dashdotted, thick] {0.5180 + 0.04957*x};
\addplot[orange!80!black, thick] {0.5624 + 0.03761*x};
\addplot[orange!50!black, loosely dotted, very thick] {0.5693 + 0.03932*x};
\addplot[blue!60!black, very thick] {0.4650 + 0.01026*x};
\addplot[black, thick] {1};
\addplot[red!70!black, thick, densely dotted] {0.27009*x};
\end{axis}
\begin{axis}[name=e, at={(d.east)}, anchor=west, xshift=0.6cm, title={Claude Haiku 5.5}, yticklabels={}, xlabel={$L/R$}]
\addplot[black!25, thin, forget plot] coordinates {(10,0) (10,1.5)};
\addplot[black!25, thin, forget plot] coordinates {(50,0) (50,1.5)};
\addplot[black!25, thin, forget plot] coordinates {(100,0) (100,1.5)};
\addplot[orange!80!black, densely dashed, thick] {0.4769 + 0.09915*x};
\addplot[gray, dotted, thick] {0.4188 + 0.06667*x};
\addplot[gray!60!black, dashed, thick] {0.6513 + 0.02222*x};
\addplot[gray!40!black, dashdotted, thick] {0.4120 + 0.04444*x};
\addplot[orange!80!black, dashdotted, thick] {0.6103 + 0.03761*x};
\addplot[orange!80!black, thick] {0.7385 + 0.01368*x};
\addplot[orange!50!black, loosely dotted, very thick] {0.6206 + 0.02222*x};
\addplot[blue!60!black, very thick] {0.5061 + 0.00684*x};
\addplot[black, thick] {1};
\addplot[red!70!black, thick, densely dotted] {0.27009*x};
\end{axis}
\begin{axis}[name=f, at={(e.east)}, anchor=west, xshift=0.6cm, title={Gemini 3.1 Flash-Lite}, yticklabels={}, xlabel={$L/R$}, legend to name=senslegend, legend columns=5, legend style={draw=none, font=\scriptsize, /tikz/every even column/.append style={column sep=6pt}}]
\addplot[black!25, thin, forget plot] coordinates {(10,0) (10,1.5)};
\addplot[black!25, thin, forget plot] coordinates {(50,0) (50,1.5)};
\addplot[black!25, thin, forget plot] coordinates {(100,0) (100,1.5)};
\addplot[orange!80!black, densely dashed, thick] {0.3539 + 0.09744*x};
\addplot[gray, dotted, thick] {0.1248 + 0.18291*x};
\addplot[gray!60!black, dashed, thick] {0.2564 + 0.12137*x};
\addplot[gray!40!black, dashdotted, thick] {0.2257 + 0.12991*x};
\addplot[orange!80!black, dashdotted, thick] {0.3898 + 0.07692*x};
\addplot[orange!80!black, thick] {0.4445 + 0.05641*x};
\addplot[orange!50!black, loosely dotted, very thick] {0.4394 + 0.05983*x};
\addplot[blue!60!black, very thick] {0.5711 + 0.01880*x};
\addplot[black, thick] {1};
\addplot[red!70!black, thick, densely dotted] {0.27009*x};
\legend{Re-solve, Generic judge, Informed judge, Step-level judge, Re-solve + generic judge, Re-solve + informed judge, Re-solve + step-level judge, \gd{}, Review everything, Accept everything}
\end{axis}
\node[anchor=north] at ($(e.south)+(0,-0.9cm)$) {\pgfplotslegendfromname{senslegend}};
\end{tikzpicture}}
\caption{Expected cost $C_{R,L}$, divided by the review cost $R$, as the cost of a false negative $L$ grows from 1 to 200 reviews (log scale). Grey verticals: the three settings of Tables~\ref{tab:detection} and~\ref{tab:detection-small}. Lower is better.}
\label{fig:cost-sensitivity}
\end{figure}

The three settings of $C_{R,L}$ in Tables~\ref{tab:detection} and~\ref{tab:detection-small} are representative examples we choose. This subsection tests whether the experiment results still hold if we alter the choice of $R$ and $L$. Since $p$ is a fixed value for each LLM choice, what actually matters is the ratio of $L/R$, i.e. how much an error matters relative to the cost of manual reviews. Figure~\ref{fig:cost-sensitivity} varies the ratio $L/R$ from 1 to 200. The expected cost is proportional to $L/R$ because $C_{R,L}/R = p/R + (n_{\mathrm{TP}}+n_{\mathrm{FP}}+n_{\mathrm{U}})/N + (L/R)\,n_{\mathrm{FN}}/N$. Reviewing everything is the flat line at 1. The expected cost of \gd{} grows the slowest among all automated methods on all models. It also intersects with the flat line the latest, meaning that \gd{} still saves money than manual review at higher $L/R$ ratio, i.e. higher-stake businees when eliminating an error matters way more than reducing cost. No baseline beats \gd{} except near the origin, where errors do not matter at all. It justifies that the comparison of Section~\ref{sec:res-detection} still holds under other choices of $R$ and $L$.

\section{Discussion and limitations}\label{sec:discussion}

\paragraph{Limitations of \gd{}}\label{sec:res-coverage}\label{sec:res-layers}\label{sec:res-readings}
\gd{} makes few wrong decisions, but its four false negatives all arise outside the kernel -- either due to the wrapper or due to corrupted data (Appendix~\ref{sec:cases}). Its main limitation is coverage: it is overly cautious and leave 78 of the 585 cases open (Table~\ref{tab:outcome-counts}). Eighteen are \outcome{Ill-posed}, mostly subjective questions such as the best place to visit, and 2 are \outcome{Contingent}. For these cases, \emph{cannot decide} is the right outcome. The other 58 are \outcome{Open}: 28 stop at the binder, which produced no valid goal; 19 stop at the kernel, which has no closure rule for the goal, such as a trip plan or the next step of a route; 7 lack a fact in the context; and 4 are undecided under some interpretations. These cases are hard for every method: the baselines give random guesses with F1 0.44--0.54.

\paragraph{Closing the coverage gap}
Closing the coverage gap requires continuing development of DATUM. The binder needs a richer binding profile, able to type-write new goals. The kernel needs a larger vocabulary of goals and closure rules, such as trip plans and route steps. Missing facts could be fetched from an external source, such as a gazetteer or a routing service, through RAG or multi-agent systems. Interpretations that leave a case undecided call for community conventions: which interpretations are admissible is a geographic question the kernel records but cannot settle \citep{scheider2023pragmatic}.

\paragraph{Toward spatial code and autonomous GIS}
This paper verifies spatial reasoning written in natural language. An autonomous GIS also writes code: queries, geoprocessing scripts and workflows whose outputs pass from one step to the next without a person in between. The same split applies there, with the wrapper transcribing code into typed statements and the kernel checking them. We aim to extend \gd{} to spatial code and to place it inside a real autonomous GIS pipeline, where it gate-keeps for each step as in our repair loop: verify, repair what is rejected, re-verify, and mitigate error accumulation along the workflows.

\section{Conclusion}\label{sec:conclusion}

AI cannot yet be trusted in geographical applications as it can in mathematics, because geographic statements rest on measurements, vague language and conventions that no formal kernel captures. We have argued that the remedy has two halves. A syntactic kernel checks typed spatial statements by rules that follow GIScience principles: it types claims by core concepts, compares measurements within their accuracy, and computes spatial relations from geometry. A semantic wrapper turns words into those statements, and declares every interpretation and convention it relies on, so that vague language is verified under each admissible interpretation rather than resolved in silence. \gd{} implements this split. Its verdicts are right far more often than those of LLM judges, its reports say, like a compiler, which step is wrong and how to fix it, and its own errors can be traced to the half that made them. Writing the kernel's rules and deciding which interpretations and conventions are admissible, and for whom, is geographers' work. As a next step, we will insert the kernel into real-world, industry autonomous GIS systems and evaluate its usability there, as the quality control of their pipelines.

\section*{Conflict of interest statement}
The author declares no conflict of interest.

\section*{Data availability statement}
The 585 verification tasks, the GPT-4o reasonings, the decision of every method on every task, and the planted-error and repair results that support the findings of this study are openly available on Zenodo at \url{https://doi.org/10.5281/zenodo.24297359}. The MapEval and GS-QA questions are available from their original publications. The source code of \gd{} is not publicly available; the prompts of its binder and transcriber are available from the author on reasonable request.

\bibliographystyle{apacite}
\bibliography{references}

\clearpage
\appendix
\section{The typed goal}\label{app:goal}
The \bnd{} returns every typed goal under one strict JSON schema of 31 fields, listed in full below; Figure~\ref{fig:binder-json} shows one filled instance. A field written \texttt{FIELD} is a pair \texttt{\{"value": string, "quote": string\}}: the bound value and the exact words of the question it was bound from. A list of words separated by \texttt{|} is the closed set of values the field may take, and \texttt{unset} stands for the empty string, which the \bnd{} writes when the question does not state the field. The \emph{envelope} selects which of the other fields the goal uses (Table~\ref{tab:goals}); fields the envelope does not use stay unset.

\begin{lstlisting}[frame=single,framerule=0.4pt,columns=fixed,basewidth=0.5em,basicstyle=\ttfamily\scriptsize,breaklines=false]
{
  "envelope": "lookup | filter | rank | geometry | route_walk | schedule |
               blocked",
  "conclusion": "unset | option | count | name | location | direction |
                 distance",
  "candidates": {
    "source": "unset | listing | options | routes | category",
    "category": {
      "key": string,
      "value": string,
      "quote": string
    }
  },
  "conditions": [{
    "kind": "within_distance | in_direction | in_region | attribute | open |
             mentions",
    "negate": boolean,
    "distance_km": FIELD,
    "minutes": FIELD,
    "metric": {"value": "unset | straight_line | network |
                travel_time", "quote": string},
    "mode": FIELD,
    "direction": FIELD,
    "toward": FIELD,
    "region": FIELD,
    "attribute": FIELD,
    "comparison": FIELD,
    "threshold": FIELD,
    "price_level": FIELD,
    "day": FIELD,
    "time": FIELD,
    "term": FIELD
  }],
  "rank_by": "unset | distance | attribute | cardinal | detour | pair_distance |
              route_minutes | route_km | feature_count | travel_minutes | area |
              length",
  "relation": "unset | direction | distance | direction_distance",
  "walk": "unset | measure | next_step | count | mentions",
  "mention": "unset | any | pass_by",
  "constraint": "unset | budget | deadline | closing",
  "reference": FIELD,
  "from": FIELD,
  "to": FIELD,
  "k": FIELD,
  "attribute": FIELD,
  "direction": FIELD,
  "day": FIELD,
  "time": FIELD,
  "route": FIELD,
  "step_anchor": FIELD,
  "feature": FIELD,
  "mode": FIELD,
  "ask": FIELD,
  "start": FIELD,
  "deadline": FIELD,
  "budget": FIELD,
  "return_mode": FIELD,
  "reference_described": {
    "key": string,
    "value": string,
    "attribute_key": string,
    "attribute_value": string,
    "place": string,
    "quote": string
  },
  "stops": [{
    "place": FIELD,
    "minutes": FIELD,
    "mode": FIELD
  }],
  "option_places": [{
    "option": integer,
    "places": [string]
  }],
  "option_counts": [{
    "option": integer,
    "count": integer
  }],
  "blocker": "unset | subjective | fact_unavailable | no_operation | other"
}

\end{lstlisting}

\section{Prompts}\label{app:prompts}
The baselines use the system prompts below, verbatim. Each call also passes the task (question, context and options), the reasoning and the conclusion statement as JSON, and fixes the reply format with a JSON schema; the step-level judge receives the reasoning split into numbered sentences by a fixed rule. The prompts call the conclusion statement the ``final answer'', and the informed judge's error kinds are the exception classes of Section~\ref{sec:method} under their code names, which end in \emph{Error}. The prompts of \gd{}'s \bnd{} and \tsb{} are available from the author on request.

\paragraph{Generic judge}
\begin{lstlisting}
You evaluate a reasoning written by another model.

You receive a task (a question, its context, and the answer options when there are any), the reasoning, and the final answer.

Is the final answer correct?

Return:
- verdict: "accept" if the final answer is correct, "reject" if it is incorrect, "cannot_tell" if you cannot decide.
- explanation: at most three sentences on why.
\end{lstlisting}

\paragraph{Informed judge}
\begin{lstlisting}
You check a spatial reasoning written by another model.

You receive the task exactly as its author saw it (the question, the context it was given, and the answer options when there are any), the author's reasoning, and the author's final answer.

Decide whether the final answer is right for the task.
- The context is the only source of facts. Coordinates, distances, travel times, opening hours, ratings, listings and route steps mean what the context writes. Do not use outside knowledge of the places.
- Option texts and the question mean what they say. Where the wording allows more than one reasonable reading, judge by the reading a careful reader would take, and say so in the explanation.
- Check the computations yourself: distances between coordinates, sums of times, clock arithmetic, counts, orderings.

Return:
- verdict: "accept" when the final answer is right, "reject" when it is wrong, "cannot_tell" when the context does not let you decide.
- first_error: when you reject, the first sentence of the reasoning where it goes wrong, copied exactly from the reasoning; otherwise "".
- error_kind: when you reject, the kind of that error, from this list; otherwise "None".
  DistanceError (a distance read or computed wrongly), TravelTimeError (a travel time read wrongly), TravelDistanceError (a travel distance read wrongly), CalculationError (arithmetic), ClockError (clock-time arithmetic), OpeningHoursError (opening hours misread), NearestError (the wrong place chosen as nearest), ExtremeError (the wrong maximum or minimum), CountError (a wrong count), DirectionError (a wrong compass direction), BearingError (a wrong angle), CoordinateError (wrong coordinates), RadiusError (inside or outside a radius misjudged), RegionError (inside or outside a region misjudged), CategoryError (a place of the wrong kind), OmittedCandidateError (a candidate left out), ContextReadError (another fact misread from the context), SelectionError (the wrong option picked from correct facts), Other.
- explanation: at most three sentences on why.
- fix: when you reject, one short instruction that would let the author correct the answer; otherwise "".
\end{lstlisting}

\paragraph{Step-level judge}
\begin{lstlisting}
You check a spatial reasoning written by another model, step by step.

You receive the task exactly as its author saw it (the question, the context it was given, and the answer options when there are any), the author's reasoning split into numbered steps, and the author's final answer.

Label every step, then decide whether the final answer is right for the task.
- The context is the only source of facts. Coordinates, distances, travel times, opening hours, ratings, listings and route steps mean what the context writes. Do not use outside knowledge of the places.
- Option texts and the question mean what they say. Where the wording allows more than one reasonable reading, judge by the reading a careful reader would take.
- Check the computations yourself: distances between coordinates, sums of times, clock arithmetic, counts, orderings.

Return:
- steps: one entry per numbered step, in order, with
  - n: the step number;
  - label: "correct" when what the step states is true given the context and the earlier correct steps, "incorrect" when it states something false or draws a wrong conclusion, "unverifiable" when it states nothing that can be checked (a plan, a restatement) or the context does not let you check it;
  - note: when incorrect, at most one sentence on what is wrong; otherwise "".
- first_incorrect: the number of the first step labelled incorrect, or 0 when none is.
- verdict: "accept" when the final answer is right, "reject" when it is wrong, "cannot_tell" when the context does not let you decide.
- explanation: at most three sentences on why.
\end{lstlisting}

\paragraph{Re-solve}
\begin{lstlisting}
You answer a spatial question.

You receive a question, its context, and the answer options when there are any, with the answer format to use.

- The context is the only source of facts. Coordinates, distances, travel times, opening hours, ratings, listings and route steps mean what the context writes. Do not use outside knowledge of the places.
- Work out what the question needs from the context: distances between coordinates, sums of times, clock arithmetic, counts, orderings.
- Give only the final answer in "answer", following native_answer_contract and wire_answer_schema exactly. Do not write out your working.
- When the context does not let you answer, choose option 0 when there are options, and null otherwise.
\end{lstlisting}

\section{Interpretation axes}\label{app:readings}
Table~\ref{tab:axes} lists the interpretation axes the wrapper declares for MapEval. Each axis applies only to the kinds of question named, and the kernel checks the goal under every combination of the axes that apply. For GS-QA the wrapper declares a single interpretation: direction sectors $45^\circ$ wide, as in the dataset's SQL templates.

\begin{table}[htbp]
\tbl{Interpretation axes declared for MapEval.}
{\small\begin{tabular}{>{\raggedright\arraybackslash}p{0.34\textwidth}>{\raggedright\arraybackslash}p{0.32\textwidth}>{\raggedright\arraybackslash}p{0.24\textwidth}}
\toprule
Open choice & Alternatives & Questions \\
\midrule
A direction word & $90^\circ$ compass sector; larger of the two coordinate offsets & direction, in a direction \\
Lying in a direction & within its sector; on its side of an axis & in a direction \\
An option number written without a unit & km; m & distance, route length, trip total \\
Nearest & by straight line; by the order of a sorted listing & nearest \\
An offered place the context does not list & not one of the nearby places; not known & nearest, within a radius, filter \\
An option that lists places & exactly the passing set; only passing places & filter \\
Several offered options pass & any of them is correct; no unique correct option & filter \\
A service the context does not list & absent; not known & place attribute \\
Travel mode, when the question names none & each mode the context lists & route \\
A road the route names twice & first mention; last mention & route step, count, mentions \\
Roundabout exits as options & exit of the next step; exit of the step at the named road & next route step \\
A roundabout in a step note & its own step; part of the step & route count, route choice \\
Whether a trip returns to its start & returns; does not return & trip order, places in a trip \\
Closing time & visit ends before closing; visit starts before closing & trip time, trip feasibility \\
An option time without AM or PM & AM; PM & trip time, itinerary \\
A time an itinerary states at a stop & arrival; departure & itinerary \\
Travel time to a reference place & to it; from it & rank by travel time \\
\bottomrule
\end{tabular}}
\label{tab:axes}
\end{table}

\section{Outcome counts}\label{app:outcomes}
Table~\ref{tab:outcome-counts} gives the outcome of every case, split by set and by whether the conclusion is right.

\begin{table}
\tbl{Outcomes by set. Each cell gives right/wrong: conclusions the published key marks right, then those it marks wrong. Wrong conclusions under \outcome{Proved} or \outcome{Confirmed} are false negatives; right conclusions under \outcome{Refuted} or \outcome{Falsified} are false positives.}
{\begin{tabular}{lrrrrrrr}
\toprule
Set & \outcome{Proved} & \outcome{Confirmed} & \outcome{Contingent} & \outcome{Refuted} & \outcome{Falsified} & \outcome{Open} & \outcome{Ill-posed} \\
\midrule
MapEval & 92/2 & 103/2 & 2/0 & 2/18 & 1/14 & 33/14 & 6/11 \\
GS-QA & 56/0 & 119/0 & 0/0 & 5/60 & 1/32 & 6/5 & 1/0 \\
\midrule
All & 148/2 & 222/2 & 2/0 & 7/78 & 2/46 & 39/19 & 7/11 \\
\bottomrule
\end{tabular}}
\label{tab:outcome-counts}
\end{table}

\section{Case studies}\label{sec:cases}\label{app:cases}

\paragraph{A right conclusion rejected over an invented tolerance (GS-QA T26/689)}
The case of Section~\ref{sec:design-example} is right: the key gives Panera Bread at 3.45 km, and the author concluded Panera Bread at 3.647 km. Both informed judges rejected it with a DistanceException, comparing 3.44 km with 3.647 km under a tolerance each chose for itself. \gd{} accepts it as \outcome{Proved} because the tolerance is a declared policy of the dataset (5\%, at least 0.5 km), the same for every method and every case. Choosing a tolerance is a decision about the result space, not about the reasoning, and a verifier should not make it silently.

\paragraph{A direction claimed without a calculation (GS-QA T4/2790)}
The question asks: \emph{What attraction can I find within 40 kilometers ahead of Memory Garden, Baltimore, MD toward National Firearms Museum, Fairfax, VA?} The question supplies the coordinates of both places and of 30 nearby attractions. A GPT-4o reasoning correctly works out that the route heads southwest, describes a filter by distance and direction, and then reports that, ``upon performing these calculations'', Dumbarton Farm Pillar lies ``in the direction towards Fairfax''. No calculation is shown, and the claim is false: the listed coordinates put Dumbarton Farm Pillar 8.6~km almost due north of the start, behind a traveller heading for Fairfax (Figure~\ref{fig:case-baltimore}). The reasoning reads well and every individual fact in it is real, which is exactly why such errors survive in an unchecked pipeline. No interpretation is at stake here: a place behind the start is not ahead of it however ``toward'' is read. \gd{} recomputes the bearing from the listed coordinates, names the step whose direction claim fails, traces it to the conclusion, computes the attractions that do qualify, and states how to fix the reasoning.

\begin{figure}[t]
\centering
\footnotesize
\setlength{\fboxsep}{4pt}
\fbox{\begin{minipage}{0.965\textwidth}
\begin{minipage}[t]{0.58\textwidth}\vspace{0pt}
\textbf{Question} (GS-QA): What attraction can I find within 40 kilometers ahead of Memory Garden, Baltimore, MD toward National Firearms Museum, Fairfax, VA?\\[3pt]
\textbf{Reasoning} (GPT-4o, abridged): \ldots\ \textbf{1. Identify the Direction}: Determine the general direction from Memory Garden, Baltimore, MD (39.30002, -76.63858) to the National Firearms Museum, Fairfax, VA (38.86295, -77.33558). This direction is approximately southwest. \ldots\ \textbf{3. Filter Attractions}: \ldots\ calculating the perpendicular distance from each place to the path \ldots\ Upon performing these calculations, ``Dumbarton Farm Pillar'' (39.37578, -76.61923) is found to be within 40 kilometers of the path \ldots\ and \hlbad{s6}{lies in the direction towards Fairfax, VA.} Therefore, it is a suitable answer to the question.\\[3pt]
\textbf{Conclusion statement}: Dumbarton Farm Pillar.
\end{minipage}\hfill
\begin{minipage}[t]{0.40\textwidth}\vspace{0pt}\centering
\begin{tikzpicture}[x=0.06cm,y=0.06cm,font=\scriptsize]
\fill[green!12] (0,0) -- (-163.8:40) arc[start angle=-163.8,end angle=-118.8,radius=40] -- cycle;
\draw[gray!60] (0,0) circle[radius=40];
\node[gray,anchor=south west,inner sep=1pt] at (45:40) {40 km};
\foreach \p in {(1.0,-1.7),(1.9,-1.0),(1.9,-1.0),(1.9,1.2),(2.0,-1.0),(-1.6,-3.1),(3.8,-0.8),(4.4,-0.9),(-4.8,0.3),(-5.0,1.8),(-5.6,1.7),(-0.4,8.9),(9.6,8.1),(4.3,15.1),(-13.4,1.7),(-13.9,-3.6),(-10.2,14.5),(4.3,-19.8),(-0.1,36.6),(9.0,-34.8)} \fill[gray!70] \p circle[radius=0.9];
\draw[-{Stealth[length=2mm]},thick,blue!60!black] (0,0) -- (-141.3:50) node[anchor=north,align=center,text width=1.6cm,yshift=-1pt] {toward Fairfax\\(77 km)};
\fill[green!50!black] (-11.1,-9.1) circle[radius=1.5] node[anchor=east,inner sep=2pt] {Tardis};
\fill[green!50!black] (-21.8,-18.3) circle[radius=1.5] node[anchor=east,inner sep=2pt] {Scott's Cove};
\fill[red!75!black] (1.7,8.4) circle[radius=1.6];
\draw[red!75!black,thin] (1.7,8.4) -- (-6,27) node[anchor=south,inner sep=1pt,align=center,text=red!75!black] {Dumbarton Farm Pillar\\(conclusion: 11$^\circ$, 8.6 km)};
\fill (0,0) circle[radius=1.6] node[anchor=north west,inner sep=1.5pt] {Memory Garden};
\draw[-{Stealth[length=1.6mm]}] (38,-50) -- (38,-40) node[anchor=south,inner sep=1pt] {N};
\end{tikzpicture}\\[-1pt]
{\scriptsize Shaded: within 40 km and within 22.5$^\circ$ of the heading to Fairfax.}
\end{minipage}
\end{minipage}}\\[5pt]
\fbox{\begin{minipage}{0.965\textwidth}
\setlength{\fboxsep}{1pt}%
\scriptsize\ttfamily\raggedright
\textbf{\gd{}}: \outcome{\rmfamily Refuted}\\[2pt]
gsqa/T4/2790: error [GD-E401]: a wrong step changes the conclusion\\
\textbf{DirectionException}: Dumbarton Farm Pillar is not southwest of Memory Garden, Baltimore, MD:\\
\hspace*{2em}its bearing is 11 deg (southwest sector 202.5 to 247.5 deg)\\
\hspace*{2em}at \ttbox{stepbad}{s6} in\symbol{95}direction(from=Memory Garden, Baltimore, MD, to=Dumbarton Farm Pillar,\\
\hspace*{4em}direction=southwest)\ \ "lies in the direction towards Fairfax, VA."\\
\hspace*{2em}at conclusion \symbol{123}"name": "Dumbarton Farm Pillar"\symbol{125}\\
\hspace*{4em}computed: Tardis, Scott's Cove at Rocky Gorge Reservoir\\
Evaluated under: direction and toward sectors are 45 degrees wide (+-22.5),\\
\hspace*{2em}as in GS-QA's SQL templates\\
\textbf{Fix}: test s6 again from the listed coordinates, then redo every step that uses it
\end{minipage}}\\[3pt]
{\scriptsize\sffamily
\colorbox{stepbad}{\strut wrong step}\quad
{\color{green!50!black}$\bullet$} valid conclusions computed by \gd{}\quad
{\color{gray!70}$\bullet$} other listed attractions}
\normalsize
\caption{A GS-QA question checked by \gd{}. Top left: the GPT-4o reasoning, with the step the kernel refuted highlighted and tagged with its step id. Top right: the listed attractions on a local kilometre grid around Memory Garden; the concluded attraction lies 8.6~km to the north, while the qualifying attractions lie in the shaded sector toward Fairfax. Bottom: the report \gd{} prints (line breaks added). The direction claim in s6 fails, the error reaches the conclusion, and the kernel computes the valid conclusions instead.}
\label{fig:case-baltimore}
\end{figure}

\paragraph{A conclusion that fails under every interpretation (MapEval 472)}
The question asks for the latest time to leave Kirstenbosch National Botanical Garden so that a visit to the Two Oceans Aquarium is still possible before it closes. The options give bare clock times, with no AM or PM. The converter therefore declares two interpretations for each option time, and the kernel checks every combination. The author's conclusion statement, 3:53 PM, is the \emph{earliest} departure after the planned stay, not the latest; it fails under every combination of interpretations, so the outcome is \outcome{Falsified} rather than \outcome{Open}. Before the half of the clock was modelled as an interpretation, this case was undecided.

\paragraph{An error in the published key (MapEval 505)}
The question asks for the direction of the pyramids of Giza from Karnak. The listed coordinates place Giza about $4.3^\circ$ north and $1.5^\circ$ west of Karnak. The author concluded North, and \gd{} \outcome{Proved} it under every declared direction interpretation. The published key says South. We count this case against \gd{} in all tables, as one of its four false negatives; with the corrected key, they fall to 3 of 374. All three baselines also accept North, so the correction lowers their false negatives by one as well.

\paragraph{A well-formed binding that names the wrong place (MapEval 58)}
The question asks for the nearest bus terminal to the Ambarkhana area of Sylhet. The context lists bus stations sorted by distance from Sylhet, not from Ambarkhana, so the published key marks the question unanswerable. The author chose the first station on the list, Airport Road Bus Stop. The binder stated the goal as the nearest station to Sylhet: a well-formed, correctly quoted binding of the wrong place. Under that goal the conclusion follows from the sorted list, and \gd{} \outcome{Proved} it. The contracts check form, not meaning, so the wrong place passed them; the report shows the bound place, where a reader can see the error.

\paragraph{A conclusion that holds under one interpretation (MapEval 246)}
The question asks whether to visit LABAID Specialized Hospital or Sultan's Dine first, leaving home at 8:30 PM by car. Both orders fit the opening hours. The author chose LABAID first, because that order finishes earlier (10:14 PM against 10:29 PM). The wrapper declares whether a trip returns to its start as an interpretation axis (Appendix~\ref{app:readings}), and the conclusion holds under only some of the declared interpretations. \gd{} accepted it with the \emph{conditional} tag, which names the interpretation a user must confirm; the published key chooses Sultan's Dine.

\paragraph{A key that disagrees with the listed coordinates (MapEval 517)}
The question asks for the second nearest park to the Tower of London. By the coordinates the context lists, Tower Bridge Piazza is nearest (about 168~m), Wakefield Gardens second (about 195~m) and Trinity Square Gardens third (about 223~m). The author computed the same order and concluded Wakefield Gardens, and \gd{} accepted it. The published key gives Trinity Square Gardens, which the listed coordinates do not support; the kernel follows the coordinates, as it must, since they are the only locations the task supplies.

\end{document}